\RequirePackage{fix-cm}
\documentclass[twocolumn]{svjour3}          
\smartqed  
\usepackage{graphicx}
\usepackage{times}
\usepackage{epsfig}
\usepackage{graphicx}
\usepackage{amsmath}
\usepackage{amssymb}
\usepackage{booktabs}
\usepackage{bm}
\usepackage{hyperref}
\usepackage{multirow}
\usepackage{multicol}
\usepackage{float}
\usepackage{stfloats}
\usepackage{subfiles}
\newcommand{\etal}{\textit{et al}. }
\newcommand{\ie}{\textit{i}.\textit{e}. }
\newcommand{\eg}{\textit{e}.\textit{g}.}
\hypersetup{
    colorlinks=true,
    linkcolor=blue,
    filecolor=magenta,      
    urlcolor=magenta,
    citecolor=blue,
}
\usepackage{natbib} 
\begin{document}

\title{FairMOT: On the Fairness of Detection and Re-Identification in Multiple Object Tracking }


\author{Yifu Zhang$^{1{\dag}}$ \and 
Chunyu Wang$^{2{\dag}}$ \and 
Xinggang Wang$^{1*}$ \and 
Wenjun Zeng$^{2}$ \and 
Wenyu Liu$^{1}$ 
}

\authorrunning{Yifu Zhang, Chunyu Wang, Xinggang Wang, Wenjun Zeng, Wenyu Liu} 

\institute{Yifu Zhang \at \email{yifuzhang@hust.edu.cn}          
          \and
           Chunyu Wang \at \email{chnuwa@microsoft.com}
           \and
           Xinggang Wang \at \email{xgwang@hust.edu.cn}
           \and
           Wenjun Zeng \at  \email{wezeng@microsoft.com}
           \and
           Wenyu Liu \at \email{liuwy@hust.edu.cn}
           \and
           $^1$\;\; Huazhong\ University\ of\ Science\ and\ Technology,\ Wuhan,\ China \\
           $^2$\;\; Microsoft\ Research\ Asia,\ Beijing,\ China \\
           $^*$\;\; Corresponding\ Author \\
           $^{\dag}$\;\; Yifu Zhang and Chunyu Wang have contributed equally. \\
}

\date{Received: date / Accepted: date}

\maketitle

\begin{abstract}

Multi-object tracking (MOT) is an important problem in computer vision which has a wide range of applications. Formulating MOT as multi-task learning of object detection and re-ID in a single network is appealing since it allows joint optimization of the two tasks and enjoys high computation efficiency. However, we find that the two tasks tend to compete with each other which need to be carefully addressed. In particular, previous works usually treat re-ID as a secondary task whose accuracy is heavily affected by the primary detection task. As a result, the network is biased to the primary detection task which is not \emph{fair} to the re-ID task.  To solve the problem, we present a simple yet effective approach termed as \emph{FairMOT} based on the anchor-free object detection architecture CenterNet. Note that it is not a naive combination of CenterNet and re-ID. Instead, we present a bunch of detailed designs which are critical to achieve good tracking results by thorough empirical studies. The resulting approach achieves high accuracy for both detection and tracking. The approach outperforms the state-of-the-art methods by a large margin on several public datasets. The source code and pre-trained models are released at \url{https://github.com/ifzhang/FairMOT}.

\end{abstract}

\section{Introduction}
Multi-Object Tracking (MOT) has been a longstanding goal in computer vision \citep{bewley2016simple,wojke2017simple,chen2018real,yu2016poi}. The goal is to estimate trajectories for objects of interest presented in videos. The successful resolution of the problem can immediately benefit many applications such as intelligent video analysis, human computer interaction, human activity recognition \citep{wang2013approach,luo2017learning}, and even social computing.

Most of the existing methods such as \citep{mahmoudi2019multi,zhou2018online,fang2018recurrent,bewley2016simple,wojke2017simple,chen2018real,yu2016poi} attempt to address the problem by two separate models: the \emph{detection} model firstly detects objects of interest by bounding boxes in each frame, then the \emph{association} model extracts re-identification (re-ID) features from the image regions corresponding to each bounding box, links the detection to one of the existing tracks or creates a new track according to certain metrics defined on features. 

There has been remarkable progress on object detection \citep{ren2015faster,he2017mask,zhou2019objects,redmon2018yolov3,fu2020model,sun2021sparse,peize2020onenet} and re-ID \citep{zheng2017person,chen2018real} respectively recently which in turn boosts the overall tracking accuracy. However, these two-step methods suffer from scalability issues. They cannot achieve real-time inference speed when there are a large number of objects in the environment because the two models do not share features and they need to apply the re-ID models for every bounding box independently in the video.

With the maturity of multi-task learning \citep{kokkinos2017ubernet,chen2018gradnorm}, one-shot trackers which estimate objects and learn re-ID features using a single network have attracted more attention \citep{wang2020towards,voigtlaender2019mots}. For example, Voigtlaender \etal \citep{voigtlaender2019mots} add a re-ID branch to Mask R-CNN to extract a re-ID feature for each proposal \citep{he2017mask}. It reduces inference time by re-using backbone features for the re-ID network. But the performance drops remarkably compared to the two-step models. In fact, the detection accuracy is still good but the tracking performance drops a lot. For example, the number of ID switches increases by a large margin. The result suggests that combining the two tasks is a non-trivial task and should be treated carefully.

In this paper, we investigate the reasons behind the failure, and present a simple yet effective solution. Three factors are identified to account for the failure. The first issue is caused by anchors. Anchors are originally designed for object detection \citep{ren2015faster}. However, we show that anchors are not suitable for extracting re-ID features for two reasons. First, anchor-based one-shot trackers such as Track R-CNN \citep{voigtlaender2019mots} overlook the re-ID task because they need anchors to first detect objects (\ie, using RPN \citep{ren2015faster}) and then extract the re-ID features based on the detection results (re-ID features are useless when detection results are incorrect). So when competition occurs between the two tasks, it will favor the detection task. Anchors also introduce a lot of ambiguity during training the re-ID features because one anchor may correspond to multiple identities and multiple anchors may correspond to one identity, especially in crowded scenes. 

The second issue is caused by feature sharing between the two tasks. Detection task and re-ID task are two totally different tasks and they need different features. In general, re-ID features need more low-level features to discriminate different instances of the same class while detection features need to be similar for different instances. The shared features in one-shot trackers will lead to feature conflict and thus reduce the performance of each task. 

The third issue is caused by feature dimension. The dimension of re-ID features is usually as high as $512$ \citep{wang2020towards} or $1024$ \citep{zheng2017person} which is much higher than that of object detection. We find that the huge difference between dimensions will harm the performance of the two tasks. More importantly, our experiments suggest that it is a generic rule that learning low-dimensional re-ID features for ``joint detection and re-ID'' networks achieves both higher tracking accuracy and efficiency. This also reveals the difference between the MOT task and the re-ID task, which is overlooked in the field of MOT.

\begin{figure*}
	\centering
	\includegraphics[width=6.5in]{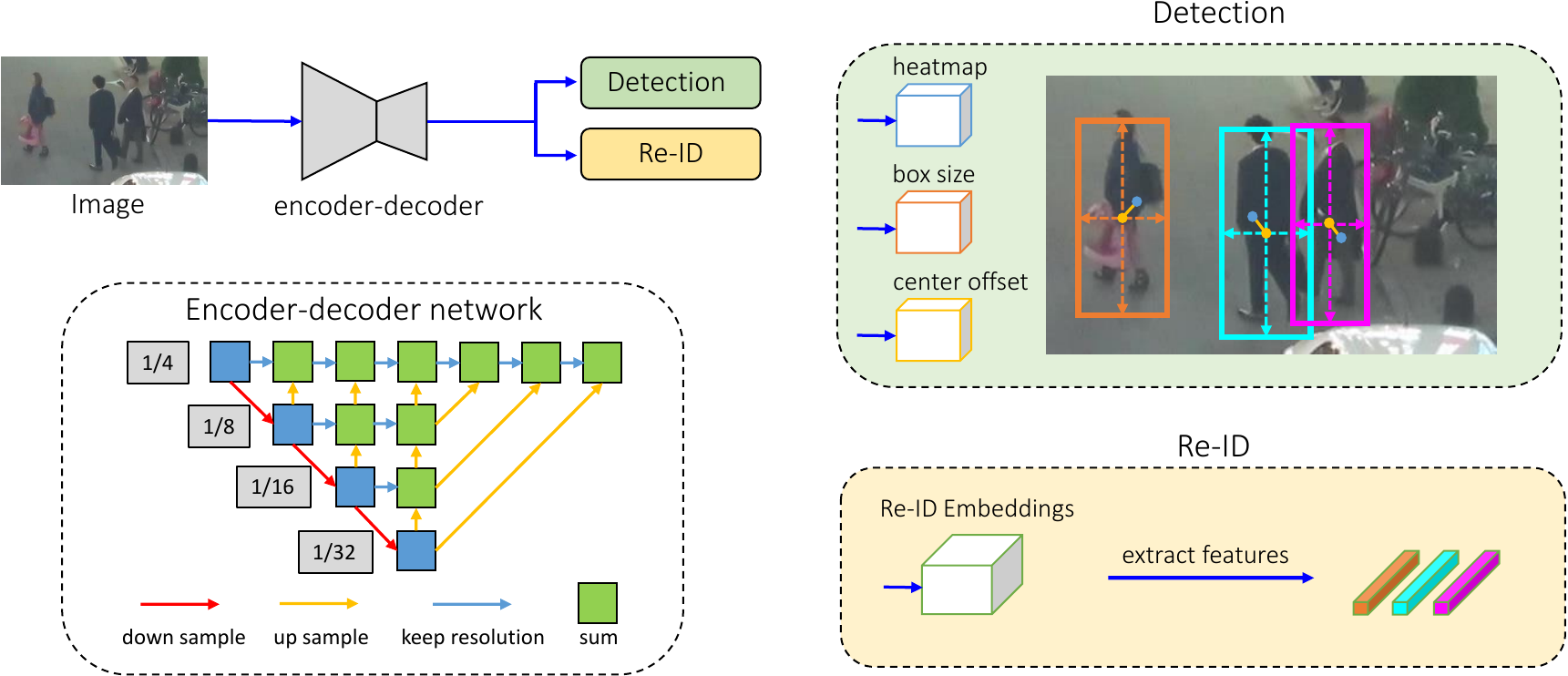}
	\caption{Overview of our one-shot tracker \emph{FairMOT}. The input image is first fed to an encoder-decoder network to extract high resolution feature maps (stride=$4$). Then we add two homogeneous branches for detecting objects and extracting re-ID features, respectively. The features at the predicted object centers are used for tracking.
	}
	\label{fig:pipeline}
\end{figure*}

In this work, we present a simple approach termed as \emph{FairMOT} which elegantly address the three issues as illustrated in Figure \ref{fig:pipeline}. \emph{FairMOT} is built on top of CenterNet \citep{zhou2019objects}. In particular, the detection and re-ID tasks are treated equally in \emph{FairMOT} which essentially differs from the previous ``detection first, re-ID secondary'' framework. It is worth noting that it is not a naive combination of CenterNet and re-ID. Instead, we present a bunch of detailed designs which are critical to achieve good tracking results by thorough empirical studies.


Figure~\ref{fig:pipeline} shows an overview of \emph{FairMOT}. It has a simple network structure which consists of two \emph{homogeneous} branches for detecting objects and extracting re-ID features, respectively. Inspired by \citep{zhou2019objects,law2018cornernet,zhou2019bottom,duan2019centernet}, the detection branch is implemented in an \emph{anchor-free} style which estimates object centers and sizes represented as position-aware measurement maps. Similarly, the re-ID branch estimates a re-ID feature for each pixel to characterize the object centered at the pixel. Note that the two branches are completely homogeneous which essentially differs from the previous methods which perform detection and re-ID in a two-stage cascaded style. So \emph{FairMOT} eliminates the unfair disadvantage of the detection branch as reflected in Table~\ref{table:anchor}, effectively learns high-quality re-ID features and obtains a good trade-off between detection and re-ID.

We evaluate \emph{FairMOT} on the MOT Challenge benchmark via the evaluation server. It ranks first among all trackers on the 2DMOT15 \citep{leal2015motchallenge}, MOT16 \citep{milan2016mot16}, MOT17 \citep{milan2016mot16} and MOT20 \citep{dendorfer2020mot20} datasets. When we further pre-train our model using our proposed single image training method, it achieves additional gains on all datasets.  In spite of the strong results, the approach is very simple and runs at 30 FPS on a single RTX 2080Ti GPU. It sheds light on the relationship between detection and re-ID in MOT and provides guidance for designing one-shot video tracking networks. 

Our contributions are as follows:
\begin{itemize}
    
    \item We empirically demonstrate that the prevalent anchor-based one-shot MOT architectures have limitations in terms of learning effective re-ID features which has been overlooked. The issues severely limit the tracking performance of those methods. 
    
    \item We present \emph{FairMOT} to address the fairness issue. \emph{FairMOT} is built on top of CenterNet. Although the adopted techniques are mostly not novel by themselves, we have new discoveries which are important to MOT. These are both novel and valuable.

    \item We show that the achieved fairness allows our \emph{FairMOT} to obtain high levels of detection and tracking accuracy and outperform the previous state-of-the-art methods by a large margin on multiple datasets such as 2DMOT15, MOT16, MOT17 and MOT20. 
    
\end{itemize}

\section{Related Work}

The best-performing MOT methods \citep{bergmann2019tracking,braso2020learning,hornakova2020lifted,yu2016poi,mahmoudi2019multi,zhou2018online,wojke2017simple,chen2018real,wang2020towards,voigtlaender2019mots,zhang2021bytetrack} usually follow the tracking-by-detection paradigm, which first detect objects in each frame and then associate them over time. We classify the existing works into two categories based on whether they use a single model or separate models to detect objects and extract association features. We discuss the pros and cons of the methods and compare them to our approach.  

\subsection{Detection and Tracking by Separate Models}

\subsubsection{Detection Methods} 
Most benchmark datasets such as MOT17 \citep{milan2016mot16} provide detection results obtained by popular methods such as DPM \citep{felzenszwalb2008discriminatively}, Faster R-CNN \citep{ren2015faster} and SDP \citep{yang2016exploit} such that the works that focus on the tracking part can be fairly compared on the same object detections. Some works such as \citep{yu2016poi,wojke2017simple,zhou2018online,mahmoudi2019multi} use a large private pedestrian detection dataset to train the  Faster R-CNN detector with VGG-16 \citep{simonyan2014very} as backbone, which obtain better detection performance. A small number of works such as \citep{han2020mat} use more powerful detectors which are developed recently such as Cascade R-CNN \citep{cai2018cascade} to boost the detection performance. 

\subsubsection{Tracking Methods}
Most of the existing works focus on the tracking part of the problem. We classify them into two classes according to the type of cues used for association. 

\paragraph{\textbf{Location and Motion Cues based Methods}}
SORT \citep{bewley2016simple} first uses Kalman Filter \citep{kalman1960new} to predict future locations of the tracklets, computes their overlap with the detections, and uses Hungarian algorithm \citep{kuhn1955hungarian} to assign detections to tracklets. IOU-Tracker \citep{bochinski2017high} directly computes the overlap between the tracklets (of the previous frame) and the detections without using using Kalman filter to predict future locations. The approach achieves $100$K fps inference speed (detection time not counted) and works well when object motion is small. Both SORT and IOU-Tracker are widely used in practice due to their simplicity. 

However, they may fail in challenging cases of crowded scenes and fast motion. Some works such as \citep{xiang2015learning,zhu2018online,chu2019famnet,chu2019online} leverage sophisticated single object tracking methods to get accurate object locations and reduce false negatives. However, these methods are extremely slow especially when there are a large number of people in the scene. To solve the problem of trajectory fragments, Zhang \etal \citep{zhang2020long} propose a motion evaluation network to learn long-range features of tracklets for association. MAT \citep{han2020mat} is an enhanced SORT, which additionally models the camera motion and uses dynamic windows for long-range re-association. 

\paragraph{\textbf{Appearance Cues based Methods}}
Some recent works \citep{yu2016poi,mahmoudi2019multi,zhou2018online,wojke2017simple} propose to crop the image regions of the detections and feed them to re-ID networks \citep{zheng2017discriminatively,hermans2017defense,luo2019bag} to extract image features. Then they compute the similarity between tracklets and detections based on re-ID features and use Hungarian algorithm \citep{kuhn1955hungarian} to accomplish assignment. The method is robust to fast motion and occlusion. In particular, it can re-initialize lost tracks because appearance features are relatively stable over time.

There are also some works \citep{bae2014robust,tang2017multiple,sadeghian2017tracking,chen2018real,xu2019spatial} focusing on enhancing appearance features. For example, Bae \etal \citep{bae2014robust} propose an online appearance learning method to handle appearance variations. Tang \etal \citep{tang2017multiple} leverage body pose features to enhance the appearance features. Some methods \citep{sadeghian2017tracking,xu2019spatial,shan2020fgagt} propose to fuse multiple cues (\ie motion, appearance and location) to get more reliable similarity. MOTDT \citep{chen2018real} proposes a hierarchical data association strategy which uses IoU to associate objects when appearance features are not reliable. A small number of works such as \citep{mahmoudi2019multi,zhou2018online,fang2018recurrent} also propose to use more complicated association strategies such as group models and RNNs.

\paragraph{\textbf{Offline Methods}}
The offline methods (or batch methods) \citep{zhang2008global,wen2014multiple,berclaz2011multiple,zamir2012gmcp,milan2013continuous,choi2015near,braso2020learning,hornakova2020lifted} often achieve better results by performing global optimization in the whole sequence. For example, Zhang \etal \citep{zhang2008global} build a graphical model with nodes representing detections in all frames. The optimal assignment is searched using a min-cost flow algorithm, which exploits the specific structure of the graph to reach the optimum faster than Linear Programming. Berclaz \etal \citep{berclaz2011multiple} also treat data association as a flow optimization task and use the K-shortest paths algorithm to solve it, which significantly speeds up computation and reduces parameters that need to be tuned. Milan \etal \citep{milan2013continuous} formulate multi-object tracking as minimization of a continuous energy and focus on designing the energy function. The energy depends on locations and motion of all targets in all frames as well as physical constraints. MPNTrack \citep{braso2020learning} proposes trainable graph neural networks to perform a global association of the entire set of detections and make MOT fully differentiable. Lif\_T \citep{hornakova2020lifted} formulates MOT as a lifted disjoint path problem and introduces lifted edges for long range temporal interactions, which significantly reduces id switches and re-identify lost.

\paragraph{\textbf{Advantages and Limitations}}
For the methods which perform detection and tracking by separate models, the main advantage is that they can develop the most suitable model for each task separately without making compromise. In addition, they can crop the image patches according to the detected bounding boxes and resize them to the same size before estimating re-ID features. This helps to handle the scale variations of objects. As a result, these approaches \citep{yu2016poi,henschel2019multiple} have achieved the best performance on the public datasets. However, they are usually very slow because the two tasks need to be done separately without sharing. So it is hard to achieve video rate inference which is required in many applications.

\subsection{Detection and Tracking by a Single Model}
With the quick maturity of multi-task learning \citep{kokkinos2017ubernet,ranjan2017hyperface,sener2018multi} in deep learning, joint detection and tracking using a single network has begun to attract more research attention. We classify them into two classes as discussed in the following.

\paragraph{\textbf{Joint Detection and Re-ID}}
The first class of methods \citep{voigtlaender2019mots,wang2020towards,liang2020rethinking,pang2021quasi,lu2020retinatrack} perform object detection and re-ID feature extraction in a single network in order to reduce inference time. For example, Track-RCNN \citep{voigtlaender2019mots} adds a re-ID head on top of Mask R-CNN \citep{he2017mask} and regresses a bounding box and a re-ID feature for each proposal. Similarly, JDE \citep{wang2020towards} is built on top of YOLOv3 \citep{redmon2018yolov3} which achieves near video rate inference. However, the accuracy of these one-shot trackers is usually lower than that of the two-step ones.

\paragraph{\textbf{Joint Detection and Motion Prediction}}
The second class of methods \citep{feichtenhofer2017detect,zhou2020tracking,pang2020tubetk,peng2020chained,sun2020transtrack} learn detection and motion features in a single network. D\&T \citep{feichtenhofer2017detect} propose a Siamese network which takes input of adjacent frames and predicts inter-frame
displacements between bounding boxes. Tracktor \citep{bergmann2019tracking} directly exploits the bounding box regression head to propagate identities of region proposals and thus removes box association. Chained-Tracker \citep{peng2020chained} proposes an end-to-end model using adjacent frame pair as input and generating the box pair representing the same target. These box-based methods assume that bounding boxes have a large overlap between frames, which is not true in low-frame rate videos. Different from these methods, CenterTrack \citep{zhou2020tracking} predicts the object center displacements with pair-wise inputs and associate by these point distances. It also provides the tracklets as an additional point-based heatmap input to the network and is then able to match objects anywhere even if the boxes have no overlap at all. However, these methods only associate objects in adjacent frames without re-initializing lost tracks and thus have difficulty handling occlusion cases. 

Our work belongs to the first class. We investigate the reasons why one-shot trackers get degraded association performance and propose a simple approach to address the problems. We show that the tracking accuracy is improved significantly without heavy engineering efforts. A concurrent work CSTrack \citep{liang2020rethinking} also aims to alleviate the conflicts between the two tasks from the perspective of features, and propose a cross-correlation network module to enable the model to learn task-dependent representations. Different from CSTrack, our method tries to address the problem from three perspectives in a systematic way and obtains notably better performances than CSTrack. CenterTrack \citep{zhou2020tracking} is also related to our work since it also uses center-based object detection framework. But CenterTrack does not extract appearance features and only links objects in adjacent frames. In contrast, \emph{FairMOT} can perform long-range association with the appearance features and handle occlusion cases. 

\paragraph{\textbf{Multi-task Learning}}
There is a large body of literature \citep{liu2019end,kendall2018multi,chen2018gradnorm,guo2018dynamic,sener2018multi} on multi-task learning which may be used to balance the object detection and re-ID feature extraction tasks. Uncertainty \citep{kendall2018multi} uses task-dependent uncertainty to automatically balance the single-task losses. MGDA is proposed in \citep{sener2018multi} to update the shared network weights by finding a common direction among the task-specific gradients. GradNorm \citep{chen2018gradnorm} controls the training of multi-task networks by simulating the task-specific gradients to be of similar magnitude. We evaluate these methods in the experimental sections. 

\subsection{Video Object Detection}
Video Object Detection (VOD) \citep{feichtenhofer2017detect,luo2019detect} is related to MOT in the sense that it leverages tracking to improve object detection performances in challenging frames. Although these methods were not evaluated on MOT datasets, some of the ideas may be valuable for the field. So we briefly review them in this section. Tang \etal \citep{tang2019object} detect object tubes in videos which aims to enhance classification scores in challenging frames based on their neighboring frames. The detection rate for small objects increases by a large margin on the benchmark dataset. Similar ideas have also been explored in \citep{han2016seq,kang2016object,kang2017object,tang2019object,pang2020tubetk}. One main limitation of these tube-based methods is that they are extremely slow especially when there are a large number of objects in videos.

\section{Unfairness Issues in One-shot Trackers}

\begin{figure*}
	\centering
	\includegraphics[width=1\linewidth]{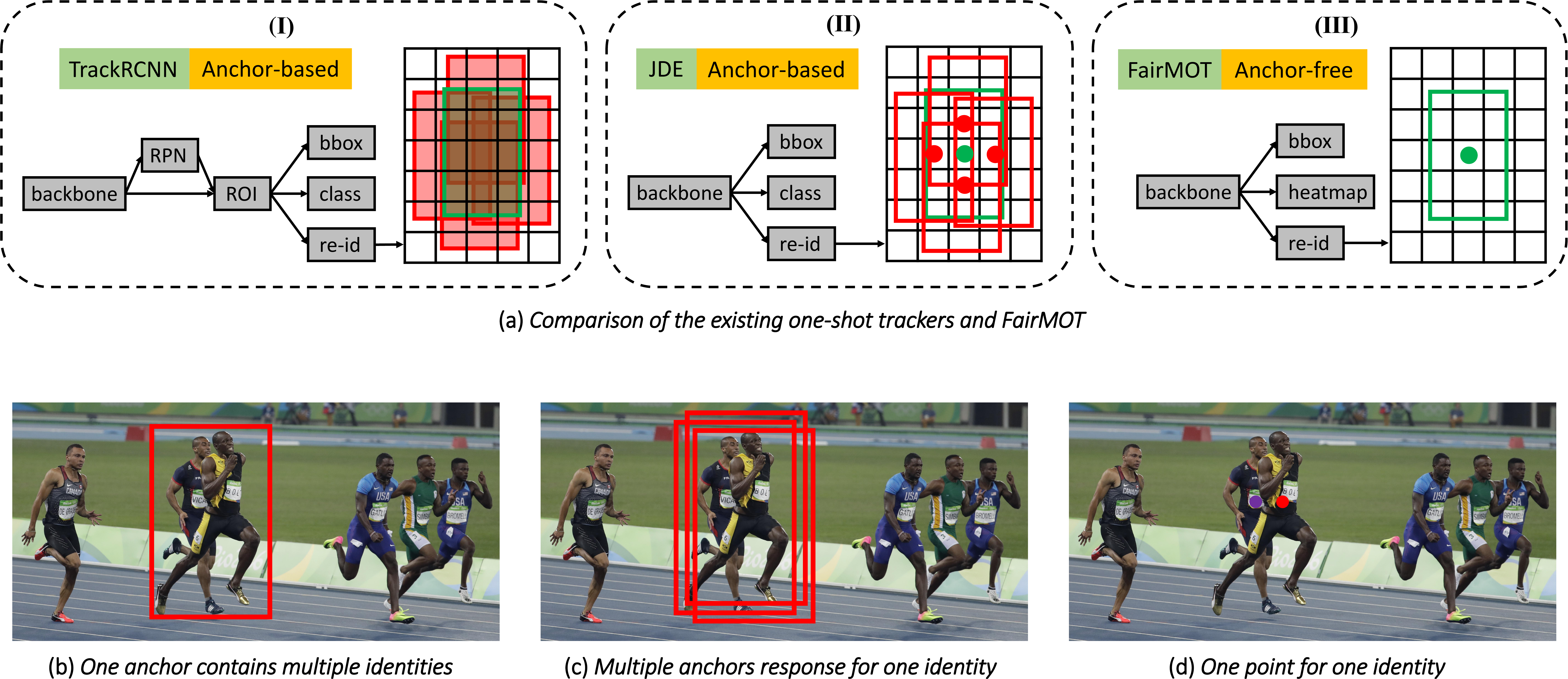}
	\caption{(a) Track R-CNN treats detection as the primary task and re-ID as the secondary one. Both Track R-CNN and JDE are anchor-based. The red boxes represent positive anchors and the green boxes represent the target objects. The three methods extract re-ID features differently. Track R-CNN extracts re-ID features for all positive anchors using ROI-Align. JDE extracts re-ID features at the centers of all positive anchors. FairMOT extracts re-ID features at the object center. (b) The red anchor contains two different instances. So it will be forced to predict two conflicting classes. (c) Three different anchors with different image patches are response for predicting the same identity. (d) FairMOT extracts re-ID features only at the object center and can mitigate the problems in (b) and (c).
	}
	\label{fig:anchor-free}
\end{figure*}

In this section, we discuss three unfairness issues that arise in the existing one-shot trackers which usually lead to degraded tracking performances. 

\subsection{Unfairness Caused by Anchors}
The existing one-shot trackers such as Track R-CNN \citep{voigtlaender2019mots} and JDE \citep{wang2020towards} are mostly anchor-based since they are directly modified from anchor-based object detectors such as YOLO \citep{redmon2018yolov3} and Mask R-CNN \citep{he2017mask}. However, we find that the anchor-based design is not suitable for learning re-ID features which result in a large number of ID switches in spite of the good detection results. We explain the problem from three perspectives in the following.

\paragraph{\textbf{Overlooked re-ID task}}
Track R-CNN \citep{voigtlaender2019mots} operates in a cascaded style which first estimates object proposals (boxes) and then pools features from them to estimate the corresponding re-ID features. The quality of re-ID features heavily depends on the quality of proposals during training (re-ID features are useless if proposals are not accurate). As a result, in the training stage, the model is seriously biased to estimate accurate object proposals rather than high quality re-ID features. So the standard ``detection first, re-ID secondary'' design of the existing one-shot trackers makes the re-ID network not fairly learned.

\paragraph{\textbf{One anchor corresponds to multiple identities}} The anchor-based methods usually use ROI-Align to extract features from proposals. Most sampling locations in ROI-Align may belong to other disturbing instances or background as shown in Figure~\ref{fig:anchor-free}. As a result, the extracted features are not optimal in terms of accurately and discriminatively representing the target objects. Instead, we find in this work that it is significantly better to only extract features at a single point, \ie, the estimated object centers. 

\paragraph{\textbf{Multiple anchors correspond to one identity}}
In both \citep{voigtlaender2019mots} and \citep{wang2020towards}, multiple adjacent anchors, which correspond to different image patches, may be forced to estimate the same identity as long as their IOU is sufficiently large. This introduces severe ambiguity for training. See Figure~\ref{fig:anchor-free} for illustration. On the other hand, when an image undergoes small perturbation, \eg, due to data augmentation, it is possible that the same anchor is forced to estimate different identities. In addition, feature maps in object detection are usually down-sampled by $8/16/32$ times to balance accuracy and speed. This is acceptable for object detection but it is too coarse for learning re-ID features because features extracted at coarse anchors may not be aligned with object centers.

\subsection{Unfairness Caused by Features}
For one-shot trackers, most features are shared between the object detection and re-ID tasks. But it is well known that they actually require features from different layers to achieve the best results. In particular, object detection requires deep features to estimate object classes and positions but re-ID requires low-level appearance features to distinguish different instances of the same class. From the perspective of the multi-task loss optimization, the optimization objectives of detection and re-ID have conflicts. Thus, it is important to balance the loss optimization strategy of the two tasks. 

\noindent
\subsection{Unfairness Caused by Feature Dimension}
The previous re-ID works usually learn very high dimensional features and have achieved promising results on the benchmarks of their field. However, we find that learning lower-dimensional features is actually better for one-shot MOT for three reasons: (1) high-dimensional re-ID features notably harms the object detection accuracy due to the competition of the two tasks which in turn also has negative impact to the final tracking accuracy. So considering that the feature dimension in object detection is usually very low (class numbers + box locations), we propose to learn low-dimensional re-ID features to balance the two tasks; (2) the MOT task is different from the re-ID task. The MOT task only performs a small number of one-to-one matchings between two consecutive frames. The re-ID task needs to match the query to a large number of candidates and thus requires more discriminative and high-dimensional re-ID features. So in MOT we do not need that high-dimensional features; (3) learning low dimensional re-ID features improves the inference speed as will be shown in our experiments. 

\section{FairMOT}

In this section, we present the technical details of \emph{FairMOT} including the backbone network, the object detection branch, the re-ID branch as well as training details. 

\subsection{Backbone Network}
We adopt ResNet-34 as backbone in order to strike a good balance between accuracy and speed. An enhanced version of Deep Layer Aggregation (DLA) \citep{zhou2019objects} is applied to the backbone to fuse multi-layer features as shown in Figure~\ref{fig:pipeline}. Different from original DLA \citep{yu2018deep}, it has more skip connections between low-level and high-level features which is similar to the Feature Pyramid Network (FPN) \citep{lin2017feature}. In addition, convolution layers in all up-sampling modules are replaced by deformable convolution such that they can dynamically adjust the receptive field according to object scales and poses. These modifications are also helpful to alleviate the alignment issue. The resulting model is named DLA-34. Denote the size of input image as $H_{\text{image}} \times W_{\text{image}}$, then the output feature map has the shape of $C \times H \times W$ where $H = H_{\text{image}} / 4$ and $W = W_{\text{image}} / 4$. Besides DLA, other deep networks that provide multi-scale convolutional features, such as Higher HRNet \citep{cheng2020bottom}, can be used in our framework to provide fair features for both detection and re-ID.

\subsection{Detection Branch}
Our detection branch is built on top of CenterNet \citep{zhou2019objects} but other anchor-free methods such as \citep{duan2019centernet,law2018cornernet,dong2020centripetalnet,yang2019reppoints} can also be used. We briefly describe the approach to make this work self-contained. In particular, three parallel heads are appended to DLA-34 to estimate heatmaps, object center offsets and bounding box sizes, respectively. Each head is implemented by applying a $3\times3$ convolution (with $256$ channels) to the output features of DLA-34, followed by a $1\times1$ convolutional layer which generates the final targets.

\subsubsection{Heatmap Head}
This head is responsible for estimating the locations of the object centers. The heatmap based representation, which is the de facto standard for the landmark point estimation task, is adopted here. In particular, the dimension of the heatmap is $1 \times H \times W$. The response at a location in the heatmap is expected to be one if it collapses with the ground-truth object center. The response decays exponentially as the distance between the heatmap location and the object center. 

For each GT box $\mathbf{b}^i = (x_1^i,y_1^i,x_2^i,y_2^i)$ in the image, we compute the object center $(c_x^i,c_y^i)$ as $c_x^i = \frac{x_1^i+x_2^i}{2}$ and $c_y^i = \frac{y_1^i+y_2^i}{2}$, respectively. Then its location on the feature map is obtained by dividing the stride $(\widetilde{c}_x^i, \widetilde{c}_y^i) = (\lfloor \frac{c_x^i}{4} \rfloor, \lfloor \frac{c_y^i}{4} \rfloor)$. Then the heatmap response at the location $(x, y)$ is computed as $M_{xy} = \sum_{i=1}^{N} \mathrm{exp}^{-\frac{(x-\widetilde{c}_x^i)^2+(y-\widetilde{c}_y^i)^2}{2\sigma _c^2}}$ where $N$ represents the number of objects in the image and $\sigma_c$ represents the standard deviation. The loss function is defined as pixel-wise logistic regression with focal loss \citep{lin2017focal}:

\begin{align}
\label{eq:heat}
\small
  L_{\text{heat}} = -\frac{1}{N} \sum _{xy} \begin{cases} (1-\hat{M}_{xy})^\alpha \text{log}(\hat{M}_{xy}), & M_{xy}=1; \\ (1-M_{xy})^\beta (\hat{M}_{xy})^\alpha \text{log}(1-\hat{M}_{xy}) & \text{otherwise},
  \end{cases}
\end{align}
where $\hat{M}$ is the estimated heatmap, and $\alpha,\beta$ are the pre-determined parameters in focal loss.

\subsubsection{Box Offset and Size Heads}
The box offset head aims to localize objects more precisely. Since the stride of the final feature map is four, it will introduce quantization errors up to four pixels. This branch estimates a continuous offset relative to the object center for each pixel in order to mitigate the impact of down-sampling. The box size head is responsible for estimating height and width of the target box at each location.

Denote the output of the \emph{size} and \emph{offset} heads as $\hat{S} \in \mathbb{R}^{2 \times H \times W}$ and $\hat{O} \in \mathbb{R}^{2 \times H \times W}$, respectively. For each GT box $\mathbf{b}^i = (x_1^i,y_1^i,x_2^i,y_2^i)$ in the image, we compute its size as $\mathbf{s}^i = (x_2^i-x_1^i, y_2^i-y_1^i)$. Similarly, the GT offset is computed as $\mathbf{o}^i = (\frac{c_x^i}{4}, \frac{c_y^i}{4}) - (\lfloor \frac{c_x^i}{4} \rfloor, \lfloor \frac{c_y^i}{4} \rfloor)$. Denote the estimated size and offset at the corresponding location as $\mathbf{\hat{s}}^i$ and $\mathbf{\hat{o}}^i$, respectively. Then we enforce $l_1$ losses for the two heads:
\begin{equation}
\label{eq:box}
    L_{\text{box}} = \sum_{i=1}^{N} \|\mathbf{o}^i - \mathbf{\hat{o}}^i\|_1 + \lambda_s \|\mathbf{s}^i - \mathbf{\hat{s}}^i\|_1.
\end{equation}
where $\lambda_s$ is a weighting parameter and is set 0.1 as the original CenterNet \citep{zhou2019objects}. 
\subsection{Re-ID Branch}

Re-ID branch aims to generate features that can distinguish objects. Ideally, affinity among different objects should be smaller than that between same objects. To achieve this goal, we apply a convolution layer with $128$ kernels on top of backbone features to extract re-ID features for each location. Denote the resulting feature map as $\mathbf{E} \in \mathbb{R}^{128 \times H \times W}$. The re-ID feature $\mathbf{E}_{x,y} \in \mathbb{R}^{128}$ of an object centered at $(x, y)$ can be extracted from the feature map. 

\subsubsection{Re-ID Loss}
We learn re-ID features through a classification task. All object instances of the same identity in the training set are treated as the same class. For each GT box $\mathbf{b}^i = (x_1^i,y_1^i,x_2^i,y_2^i)$ in the image, we obtain the object center on the heatmap $(\widetilde{c}_x^i, \widetilde{c}_y^i)$. We extract the re-ID feature vector $\mathbf{E}_{\widetilde{c}_x^i, \widetilde{c}_y^i}$ and use a fully connected layer and a softmax operation to map it to a class distribution vector $\mathbf{P} = \{\mathbf{p}(k), k \in [1,K] \}$. Denote the one-hot representation of the GT class label as $\mathbf{L}^i{(k)}$. Then we compute the re-ID loss as:
\begin{equation}
\label{eq:id}
    L_{\text{identity}} = - \sum_{i=1}^{N} \sum_{k=1}^{K} \mathbf{L}^i{(k)} \text{log}(\mathbf{p}(k)),
\end{equation}
where $K$ is the number of all the identities in the training data. During the training process of our network, only the identity embedding vectors located at object centers are used for training, since we can obtain object centers from the objectness heatmap in testing.

\subsection{Training FairMOT}
\label{sec:training}
We jointly train the detection and re-ID branches by adding the losses (i.e., Eq.~\eqref{eq:heat}, Eq.~\eqref{eq:box} and Eq.~\eqref{eq:id}) together. In particular, we use the uncertainty loss proposed in \citep{kendall2018multi} to automatically balance the detection and re-ID tasks: 
\begin{equation}
\label{eq:det}
    L_{\text{detection}} = L_{\text{heat}} + L_{\text{box}},
\end{equation}
\begin{equation}
\label{eq:uncertainty}
    L_{\text{total}} = \frac{1}{2}(\frac{1}{e^{w_1}}L_{\text{detection}} + \frac{1}{e^{w_2}}L_{\text{identity}} + w_1 + w_2),
\end{equation}
where $w_1$ and $w_2$ are learnable parameters that balance the two tasks. Specifically, given an image with a few objects and their corresponding IDs, we generate heatmaps, box offset and size maps as well as one-hot class representation of the objects. These are compared to the estimated measures to obtain losses to train the whole network. 

In addition to the standard training strategy presented above, we propose a single image training method to train \emph{FairMOT} on image-level object detection datasets such as COCO \citep{lin2014microsoft} and CrowdHuman \citep{shao2018crowdhuman}. Different from CenterTrack \citep{zhou2020tracking} that takes two simulated consecutive frames as input, we only take a single image as input. We assign each bounding box a unique identity and thus regard each object instance in the dataset as a separate class. We apply different transformations to the whole image including HSV augmentation, rotation, scaling, translation and shearing. The single image training method has significant empirical values. First, the pre-trained model on the CrowdHuman dataset can be directly used as a tracker and get acceptable results on MOT datasets such as MOT17 \citep{milan2016mot16}. This is because the CrowdHuman dataset can boost the human detection performance and also has strong domain generalization ability. Our training of the re-ID features further enhances the association ability of the tracker. Second, we can finetune it on other MOT datasets and further improve the final performance. 

\subsection{Online Inference}
In this section, we present how we perform online inference, and in particular, how we perform association with the detections and re-ID features.

\subsubsection{Network Inference}
The network takes a frame of size $1088 \times 608$ as input which is the same as the previous work JDE \citep{wang2020towards}. On top of the predicted heatmap, we perform non-maximum suppression (NMS) based on the heatmap scores to extract the peak keypoints. The NMS is implemented by a simple $3 \times 3$ max pooling operation as in \citep{zhou2019objects}. We keep the locations of the keypoints whose heatmap scores are larger than a threshold. Then, we compute the corresponding bounding boxes based on the estimated offsets and box sizes. We also extract the identity embeddings at the estimated object centers. In the next section, we discuss how we associate the detected boxes over time using the re-ID features.

\subsubsection{Online Association}
We follow MOTDT \citep{chen2018real} and use a hierarchical online data association method. We first initialize a number of tracklets based on the detected boxes in the first frame. Then in the subsequent frame, we link the detected boxes to the existing tracklets using a two-stage matching strategy. In the first stage, we use Kalman Filter \citep{kalman1960new} and re-ID features to obtain initial tracking results. In particular, we use Kalman Filter to predict tracklet locations in the following frame and compute the Mahalanobis distance $D_m$ between the predicted and detected boxes following DeepSORT \citep{wojke2017simple}. We fuse the Mahalanobis distance with the cosine distance computed on re-ID features:  $D = \lambda D_r + (1 - \lambda) D_m$ where $\lambda$ is a weighting parameter and is set to be $0.98$ in our experiments. Following JDE \citep{wang2020towards}, we set Mahalanobis distance to infinity if it is larger than a threshold to avoid getting trajectories with large motion. We use Hungarian algorithm \citep{kuhn1955hungarian} with a matching threshold $\tau_1 = 0.4$ to complete the first stage matching.

In the second stage, for unmatched detections and tracklets, we try to match them according to the overlap between their boxes. In particular, we set the matching threshold $\tau_2 = 0.5$. We update the appearance features of the tracklets in each time step to handle appearance variations as in \citep{bolme2010visual,henriques2014high}. Finally, we initialize the unmatched detections as new tracks and save the unmatched tracklets for 30 frames in case they reappear in the future.

\section{Experiments}
\subsection{Datasets and Metrics}
\label{sec:dataset}

There are six training datasets briefly introduced as follows: the ETH \citep{ess2008mobile} and CityPerson \citep{zhang2017citypersons} datasets only provide box annotations so we only train the detection branch on them. The CalTech \citep{dollar2009pedestrian}, MOT17 \citep{milan2016mot16}, CUHK-SYSU \citep{xiao2017joint} and PRW \citep{zheng2017person} datasets provide both box and identity annotations which allows us to train both branches. Some videos in ETH also appear in the testing set of the MOT17 which are removed from the training dataset for fair comparison. The overall training strategy is described in Section~\ref{sec:training}, which is the same as \citep{wang2020towards}. For the self-supervised training of our method, we use the CrowdHuman dataset \citep{shao2018crowdhuman}  which only contains object bounding box annotations.

We evaluate our approach on the testing sets of four benchmarks: 2DMOT15, MOT16, MOT17 and MOT20. We use Average Precision (\emph{AP}) to evaluate detection results. Following \citep{wang2020towards}, we use True Positive Rate (\emph{TPR}) at a false accept rate of $0.1$ for evaluating re-ID features. In particular, we extract re-ID features which correspond to ground truth boxes and use each feature to retrieve $N$ most similar candidates. We report the true positive rate at false accept rate 0.1 (\emph{TPR}@\emph{FAR}=0.1). Note that \emph{TPR} is not affected by detection results and faithfully reflects the quality of re-ID features. We use the CLEAR metric \citep{bernardin2008evaluating} (\ie \emph{MOTA}, \emph{IDs}) and \emph{IDF1} \citep{ristani2016performance} to evaluate overall tracking accuracy.

\subsection{Implementation Details}
We use a variant of DLA-34 proposed in \citep{zhou2019objects} as our default backbone. The model parameters pre-trained on the COCO dataset \citep{lin2014microsoft} are used to initialize our model. We train our model with the Adam optimizer \citep{kingma2014adam} for $30$ epochs with a starting learning rate of $10^{-4}$. The learning rate decays to $10^{-5}$ at $20$ epochs. The batch size is set to be $12$. We use standard data augmentation techniques including rotation, scaling and color jittering. The input image is resized to $1088\times 608$ and the feature map resolution is $272 \times 152$. The training step takes about 30 hours on two RTX 2080 Ti GPUs. 

\begin{table}
\begin{center}
\setlength{\tabcolsep}{3pt}
\caption{Comparison of different re-ID feature extraction (sampling) strategies on the validation set of MOT17. The rest of the models are kept the same for fair comparison. $\uparrow$ means the larger the better and $\downarrow$ means the smaller the better. The best results are shown in {\bf bold}.}
\label{table:anchor}
\begin{tabular}{lcccccc}
\toprule
Feature Extraction & Anchor & MOTA$\uparrow$ & IDF1$\uparrow$ & IDs$\downarrow$ & TPR$\uparrow$\\
\midrule
FairMOT (ROI-Align) & \checkmark & 68.7 & 71.0 & 331 & 93.1\\
FairMOT (POS-Anchor) & \checkmark & 69.0 & 70.3 & 434 & 93.9\\
FairMOT (Center) & & \textbf{69.1} & 72.8 & \textbf{299} & 94.4\\
FairMOT (Center-BI) & & 68.8 & \textbf{74.3} & 303 & \textbf{94.9}\\
FairMOT (Two-Stage) & \checkmark & 69.0 & 68.2 & 388 & 90.5\\
\bottomrule
\end{tabular}
\end{center}
\end{table}

\subsection{Ablative Studies}
In this section, we present rigorous studies of the three critical factors in \emph{FairMOT} including anchor-less re-ID feature extraction, feature fusion and feature dimensions by carefully designing a number of baseline methods.

\subsubsection{Anchors}
\label{sec:anchor}

We evaluate four strategies for sampling re-ID features from the detected boxes which are frequently used by previous works \citep{wang2020towards} \citep{voigtlaender2019mots}. The first strategy is ROI-Align used in Track R-CNN \citep{voigtlaender2019mots}. It samples features from the detected proposals using ROI-Align. As discussed previously, many sampling locations deviate from object centers. The second strategy is POS-Anchor used in JDE \citep{wang2020towards}. It samples features from positive anchors which may also deviate from object centers. The third strategy is ``Center'' used in \emph{FairMOT}. It only samples features at object centers. Recall that, in our approach, re-ID features are extracted from discretized low-resolution maps. In order to sample features at accurate object locations, we also try to apply Bi-linear Interpolation (Center-BI) to extract more accurate features. 

We also evaluate a two-stage approach to first detect object bounding boxes and then extract re-ID features. In the first stage, the detection part is the same as our \emph{FairMOT}. In the second stage, we use ROI-Align \citep{he2017mask} to extract the backbone features based on the detected bounding boxes and then use a re-ID head (a fully connected layer) to get re-ID features. The main difference between the two-stage approach and the one-stage ``ROI-Align'' approach is that the re-ID features of the two-stage approach rely on the detection results while those of the one-stage approach do not during training.

The results are shown in Table~\ref{table:anchor}. Note that the five approaches are all built on top of \emph{FairMOT}. The only difference lies in how they sample re-ID features from detected boxes. First, we can see that our approach (Center) obtains notably higher \emph{IDF1} score and True Positive Rate (\emph{TPR}) than ROI-Align, POS-Anchor and the two-stage approach. This metric is independent of object detection results and faithfully reflects the quality of re-ID features. In addition, the number of ID switches (\emph{IDs}) of our approach is also significantly smaller than the two baselines. The results validate that sampling features at object centers is more effective than the strategies used in the previous works. Bi-linear Interpolation (Center-BI) achieves even higher \emph{TPR} than Center because it samples features at more accurate locations. The two-stage approach harms the quality of the re-ID features.

\subsubsection{Balancing Multi-task Losses}
\label{sec:multi-task}
We evaluate different methods for balancing the losses of different tasks including Uncertainty \citep{kendall2018multi}, GradNorm \citep{chen2018gradnorm} and MGDA-UB \citep{sener2018multi}. We also evaluate a baseline with fixed weights obtained by grid search. We implement two versions for the uncertainty-based method. The first is ``Uncertainty-task'' which learns two parameters for the detection loss and re-ID loss, respectively. The second is ``Uncertainty-branch'' which learns four parameters for the heatmap loss, box size loss, offset loss and re-ID losses, respectively.

\begin{table}[h!]
\begin{center}
\setlength{\tabcolsep}{5pt}
\caption{Comparison of different loss weighting strategies on the validation set of the MOT17 dataset. The best results are shown in {\bf bold}.}
\label{table:loss-weighting}
\begin{tabular}{lccccc}
\toprule
Loss Weighting & MOTA$\uparrow$ & IDF1$\uparrow$ & IDs$\downarrow$ & AP$\uparrow$ & TPR$\uparrow$\\
\midrule
Fixed & \textbf{69.6} & 71.6 & 387 & \textbf{81.9} & 93.8\\
Uncertainty-task & 69.1 & 72.8 & \textbf{299} & 81.2 & 94.4\\
Uncertainty-branch & 68.5 & 73.3 & 319 & 81.0 & 96.0\\
MGDA-UB & 63.6 & 67.9 & 355 & 78.5 & \textbf{97.0}\\
GradNorm & 69.5 & \textbf{73.8} & 311 & 81.3 & 95.1\\
\bottomrule
\end{tabular}
\end{center}
\end{table}

\begin{table}[h!]
\begin{center}
\setlength{\tabcolsep}{1.5pt}
\caption{Comparison of different backbones on the validation set of MOT17 dataset. ``MLFF'' is short for multi-layer feature fusion. ``Acc'' is short for ImageNet classification accuracy. The results of the ImageNet classification accuracy are from the original papers of the backbone networks. The best results are shown in {\bf bold}. }
\label{table:backbone}
\resizebox{1\linewidth}{!}{
\begin{tabular}{lccccccc}
\toprule
Backbone & w/ MLFF & MOTA$\uparrow$ & IDF1$\uparrow$ & IDs$\downarrow$ & AP$\uparrow$ & TPR$\uparrow$ & Acc$\uparrow$\\
\midrule
ResNet-34 &  & 63.6 & 67.2 & 435 & 75.1 & 90.9 & 75.2\\
ResNet-50 &  & 63.7 & 67.7 & 501 & 75.5 & 91.9 & 77.8\\
RegNetY-4.0GF &  & 63.9 & 68.0 & 407 & 75.8 & 91.9 & \textbf{79.4}\\
ResNet-34-FPN & $\checkmark$ & 64.4 & 69.6 & 369 & 77.7 & 94.2 & 75.2\\
RegNetY-4.0GF-FPN & $\checkmark$ & 65.8 & 69.3 & 257 & 78.0 & 94.3 & \textbf{79.4}\\
HRNet-W18 & $\checkmark$ & 67.4 & 74.3 & 315 & 80.5 & 94.6 & 76.8\\
DLA-34 & $\checkmark$ & 69.1 & 72.8 & 299 & 81.2 & 94.4 & 76.9\\
HarDNet-85 & $\checkmark$ & \textbf{71.2} & \textbf{74.5} & \textbf{198} & \textbf{82.6} & \textbf{95.8} & 77.0\\
\bottomrule
\end{tabular}
}
\end{center}
\end{table}

The results are shown in Table \ref{table:loss-weighting}. We can see that the ``Fixed'' method gets the best \emph{MOTA} and \emph{AP} but the worst \emph{IDs} and \emph{TPR}. It means that the model is biased to the detection task. MGDA-UB has the highest \emph{TPR} but the lowest \emph{MOTA} and \emph{AP}, which indicates that the model is biased to the re-ID task. Similar results can be found in \citep{wang2020towards,vandenhende2021multi}. GradNorm gets the best overall tracking accuracy (highest \emph{IDF1} and second highest \emph{MOTA}), meaning that ensuring different tasks to have similar gradient magnitude is helpful to handle feature conflicts. However, GradNorm takes longer training time. So we use the simpler Uncertainty method which is slightly worse than GradNorm in the rest of our experiments. 

\subsubsection{Multi-layer Feature Fusion}
\label{sec:backbone}

We compare a number of backbones such as vanilla ResNet \citep{he2016deep}, Feature Pyramid Network (FPN) \citep{lin2017feature}, High-Resolution Network (HRNet) \citep{wang2020deep}, DLA \citep{zhou2019objects}, HarDNet \citep{chao2019hardnet} and RegNet \citep{radosavovic2020designing}. Note that the rest of the factors of these approaches such as training datasets are all controlled to be the same for fair comparison. In particular, the stride of the final feature map is four for all methods. We add three up-sampling operations for vanilla ResNet and RegNet to obtain feature maps of stride four. We divide these backbones into two categories, one without multi-layer fusion (\ie ResNet and RegNet) and one with (\ie FPN, HRNet, DLA and HarDNet).

The results are shown in Table~\ref{table:backbone}. We also list the ImageNet \citep{russakovsky2015imagenet} classification accuracy \emph{Acc} in order to demonstrate that a strong backbone in one task does not mean it will also get good results in MOT. So detailed studies for MOT are necessary and useful.

By comparing the results of ResNet-34 and ResNet-50, we find that blindly using a larger network does not notably improve the overall tracking result measured by \emph{MOTA}. In particular, the quality of re-ID features barely benefits from the larger network. For example, \emph{IDF1} only improves from $67.2\%$ to $67.7\%$ and \emph{TPR} improves from $90.9\%$ to $91.9\%$, respectively. In addition, the number of \emph{ID switches} even increases from $435$ to $501$. By comparing ResNet-50 and RegNetY-4.0GF, we can find that using a even more powerful backbone also achieves very limited gain. The re-ID metric \emph{TPR} of RegNetY-4.0GF is the same as ResNet-50 (91.9) while the ImageNet classification accuracy improves a lot (79.4 vs 77.8). All these results suggest that directly using a larger or a more powerful network cannot always improve the final tracking accuracy.

In contrast, ResNet-34-FPN, which actually has fewer parameters than ResNet-50, achieves a larger \emph{MOTA} score than ResNet-50. More importantly, \emph{TPR} improves significantly from $90.9\%$ to $94.2\%$. By comparing RegNetY-4.0GF-FPN with RegNetY-4.0GF, we can see that adding multi-layer feature fusion structure \citep{lin2017feature} to RegNet brings considerable gains (+1.9 MOTA, +1.3 IDF1, -36.9\% IDs, +2.2 AP, +2.3 TPR), which suggests that multi-layer feature fusion has clear advantages over simply using larger or more powerful networks. 

In addition, DLA-34, which is also built on top of ResNet-34 but has more levels of feature fusion, achieves an even larger \emph{MOTA} score. In particular, \emph{TPR} increases significantly from $90.9\%$ to $94.4\%$ which in turn decreases the number of ID switches (\emph{IDs}) from $435$ to $299$. Similar conclusions can be obtained from the results of HRNet-W18. The results validate that feature fusion (FPN, DLA and HRNet) effectively improves the discriminative ability of re-ID features. On the other hand, although ResNet-34-FPN obtains equally good re-ID features (\emph{TPR}) as DLA-34, its detection results (\emph{AP}) are significantly worse than DLA-34. We think the use of deformable convolution in DLA-34 is the main reason because it enables more flexible receptive fields for objects of different sizes - it is very important for our method since \emph{FairMOT} only extracts features from object centers without using any region features. We can only get 65.0 \emph{MOTA} and 78.1 \emph{AP} when replacing all the deformable convolutions with normal convolutions in DLA-34. As shown in Table~\ref{table:backbone_size}, we can see that DLA-34 mainly outperforms HRNet-W18 on middle and large size objects. When we further use a more powerful backbone HarDNet-85 with more multi-layer feature fusion structures, we achieve even better results than DLA-34 (+2.1 MOTA, +1.7 IDF1, -33.8\% IDs, +1.4 AP, +1.4 TPR). Although HRNet-W18, DLA-34 and HarDNet-85 get lower ImageNet classification accuracy than ResNet-50 and RegNetY-4.0GF, they achieve much higher tracking accuracy. Based on the experimental results above, we believe that multi-layer feature fusion is the key to solve the ``feature'' issue. 

\begin{table}
\begin{center}
\setlength{\tabcolsep}{5pt}
\caption{Demonstration of \emph{feature conflict} between the detection and re-ID tasks on the validation set of the MOT17 dataset. ``-det'' means only the detection branch is trained and the re-ID branch is randomly initialized. The best results are shown in {\bf bold}.}
\label{table:det-only}
\begin{tabular}{lccccc}
\toprule
Backbone & MOTA$\uparrow$ & IDF1$\uparrow$ & IDs$\downarrow$ & AP$\uparrow$ & TPR$\uparrow$\\
\midrule
ResNet-34 & 63.6 & 67.2 & 435 & 75.1 & 90.9\\
ResNet-34-det & 63.7 & 60.4 & 597 & 76.1 & 36.7\\
DLA-34 & \textbf{69.1} & \textbf{72.8} & \textbf{299} & \textbf{81.2} & \textbf{94.4}\\
\bottomrule
\end{tabular}
\end{center}
\end{table}

\begin{table}
\begin{center}
\setlength{\tabcolsep}{0.5pt}
\caption{The impact of different backbones on objects of different scales. \emph{Small}: area smaller than 7000 pixels; \emph{Medium}: area from 7000 to 15000 pixels; \emph{Large}: area larger than 15000 pixels. The best results are shown in {\bf bold}.}
\label{table:backbone_size}
\begin{tabular}{l|ccc|ccc|ccc}
\toprule
Backbone & AP$^S$ & AP$^M$ & AP$^L$ & TPR$^S$ & TPR$^M$ & TPR$^L$ & IDs$^S$ & IDs$^M$ & IDs$^L$\\
\midrule
ResNet-34 & 40.6 & 57.8 & 85.2 & 91.7 & 85.7 & 88.8 & 190 & 87 & 118\\
ResNet-50 & 39.7 & 59.4 & 86.0 & 91.3 & 85.3 & 89.0 & 248 & 91 & 124\\
ResNet-34-FPN & 45.9 & 61.0 & 85.4 & 90.7 & 91.5 & \textbf{93.3} & 166 & 71 & 90\\
HRNet-W18 & \textbf{51.1} & 63.7 & 85.7 & \textbf{94.2} & \textbf{92.5} & 93.1 & 168 & \textbf{55} & \textbf{56}\\
DLA-34 & 46.8 & \textbf{65.1} & \textbf{88.8} & 92.7 & 91.2 & 91.8 & \textbf{134} & 64 & 70\\
\bottomrule
\end{tabular}
\end{center}
\end{table}

\begin{table}[h!]
\begin{center}
\setlength{\tabcolsep}{4pt}
\caption{Evaluation of re-ID feature dimensions of JDE and FairMOT on the validation set of MOT17. The best results of the same method are shown in {\bf bold}. }
\label{table:dimension}
\begin{tabular}{lccccccc}
\toprule
Method & Dim & MOTA $\uparrow$ & IDF1 $\uparrow$ & IDs $\downarrow$ & AP$\uparrow$ & TPR $\uparrow$ & FPS$ \uparrow$\\
\midrule
JDE & 512 & 59.9 & 64.1 & 536 & 73.3 & 76.8 & 22.2\\
JDE & 64 & \textbf{60.3} & \textbf{65.0} & \textbf{474} & \textbf{73.6} & \textbf{82.0} & \textbf{24.4}\\
\midrule
FairMOT & 512 & 68.5 & \textbf{73.7} & 312 & 80.9 & \textbf{94.6} & 24.1\\
FairMOT & 64 & \textbf{69.2} & 73.3 & \textbf{283} & \textbf{81.3} & 94.3 & \textbf{26.8}\\
\bottomrule
\end{tabular}
\end{center}
\end{table}

To validate the existence of \emph{feature conflict} between the detection and re-ID tasks, we introduce a baseline ResNet-34-det which only trains the detection branch (re-ID branch is randomly initialized). We can see from Table~\ref{table:det-only} that the detection result measured by \emph{AP} improves by 1 point if we do not train the re-ID branch which shows the conflict between the two tasks. In particular, ResNet-34-det even gets higher \emph{MOTA} score than ResNet-34 because the metric favors better detection than tracking results. In contrast, DLA-34, which adds multi-layer feature fusion over ResNet-34, achieves better detection as well as tracking results. It means multi-layer feature fusion helps alleviate the \emph{feature conflict} problem by allowing each task to extract whatever it needs for its own task from the fused features.

\begin{figure*}
	\centering
	\includegraphics[width=6.5in]{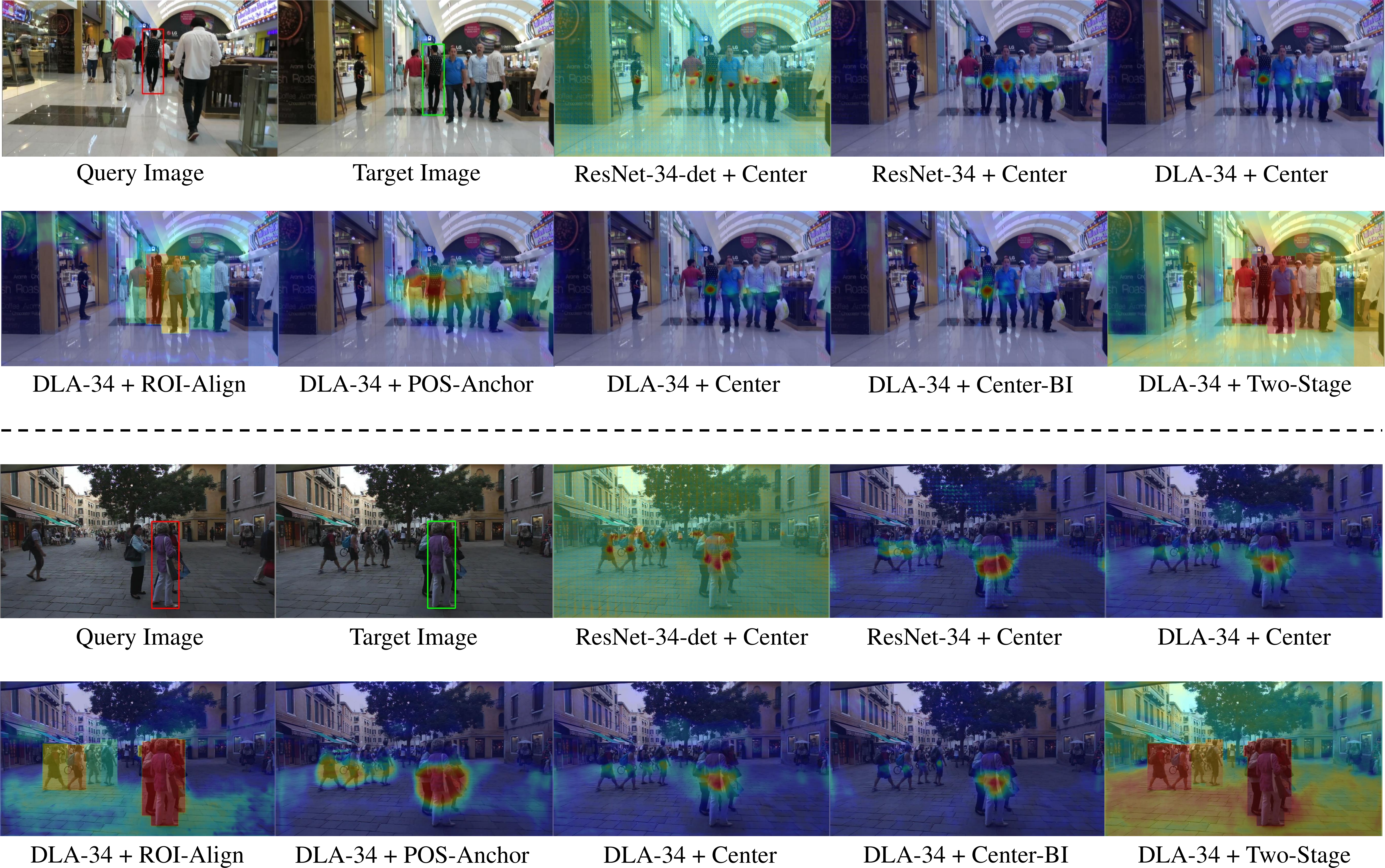}
	\caption{Visualization of the discriminative ability of the re-ID features. Query instances are marked as red boxes and target instances are marked as green boxes. The similarity maps are computed using re-ID features extracted based on different strategies (\eg, Center, Center-BI, ROI-Align and POS-Anchor as described in Section~\ref{sec:anchor}) and different backbones (\eg, ResNet-34 and DLA-34). The query frames and target frames are randomly chosen from the MOT17-09 and the MOT17-02 sequence. 
	}
	\label{fig:emb}
\end{figure*}

\subsubsection{Feature Dimension} 
The previous one-shot trackers such as JDE \citep{wang2020towards} usually learns $512$ dimensional re-ID features following the two-step methods without ablation study. However, we find in our experiments that the feature dimension actually plays an important role in balancing detection and tracking accuracy. Learning lower dimensional re-ID features causes less harm to the detection accuracy and improves the inference speed. We conduct experiments on different one-shot trackers and find it is a generic rule that low-dimensional (\ie 64) re-ID features achieves better performance than high-dimensional (\ie 512) re-ID features. 

We evaluate multiple choices for re-ID feature dimension of JDE and \emph{FairMOT} in Table~\ref{table:dimension}. For JDE, we can see that $64$ achieves better performance than $512$ on all the metrics. For \emph{FairMOT}, we can see that $512$ achieves higher \emph{IDF1} and \emph{TPR} scores which indicates that higher dimensional re-ID features lead to stronger discriminative ability. However, the \emph{MOTA} score improves when we decrease the dimension from $512$ to $64$. This is mainly caused by the conflict between the detection and re-ID tasks. In particular, we can see that the detection result (\emph{AP}) improves when we decrease the dimension of re-ID features. Different from the re-ID task, low-dimensional re-ID features achieves better performance and efficiency on the MOT task.

\subsubsection{Data Association Methods} 
This section evaluates the three ingredients in the data association step including bounding box IoU, re-ID features and Kalman Filter \citep{kalman1960new}. These are used to compute the similarity between each pair of detected boxes. With that we use Hungarian algorithm \citep{kuhn1955hungarian} to solve the assignment problem. Table~\ref{table:association} shows the results. We can see that only using box IoU causes a lot of \emph{ID switches}. This is particularly true for crowded scenes and fast camera motion. Using re-ID features alone notably increases \emph{IDF1} and decreases the number of \emph{ID switches}. In addition, adding Kalman filter helps obtain smooth (reasonable) tracklets which further decreases the number of \emph{ID switches}. When an object is partly occluded, its re-ID features become unreliable. In this case, it is important to leverage box IoU, re-ID features and Kalman filter to obtain good tracking performance. 

\begin{figure}[h!]
\centering
\includegraphics[width=0.95\linewidth]{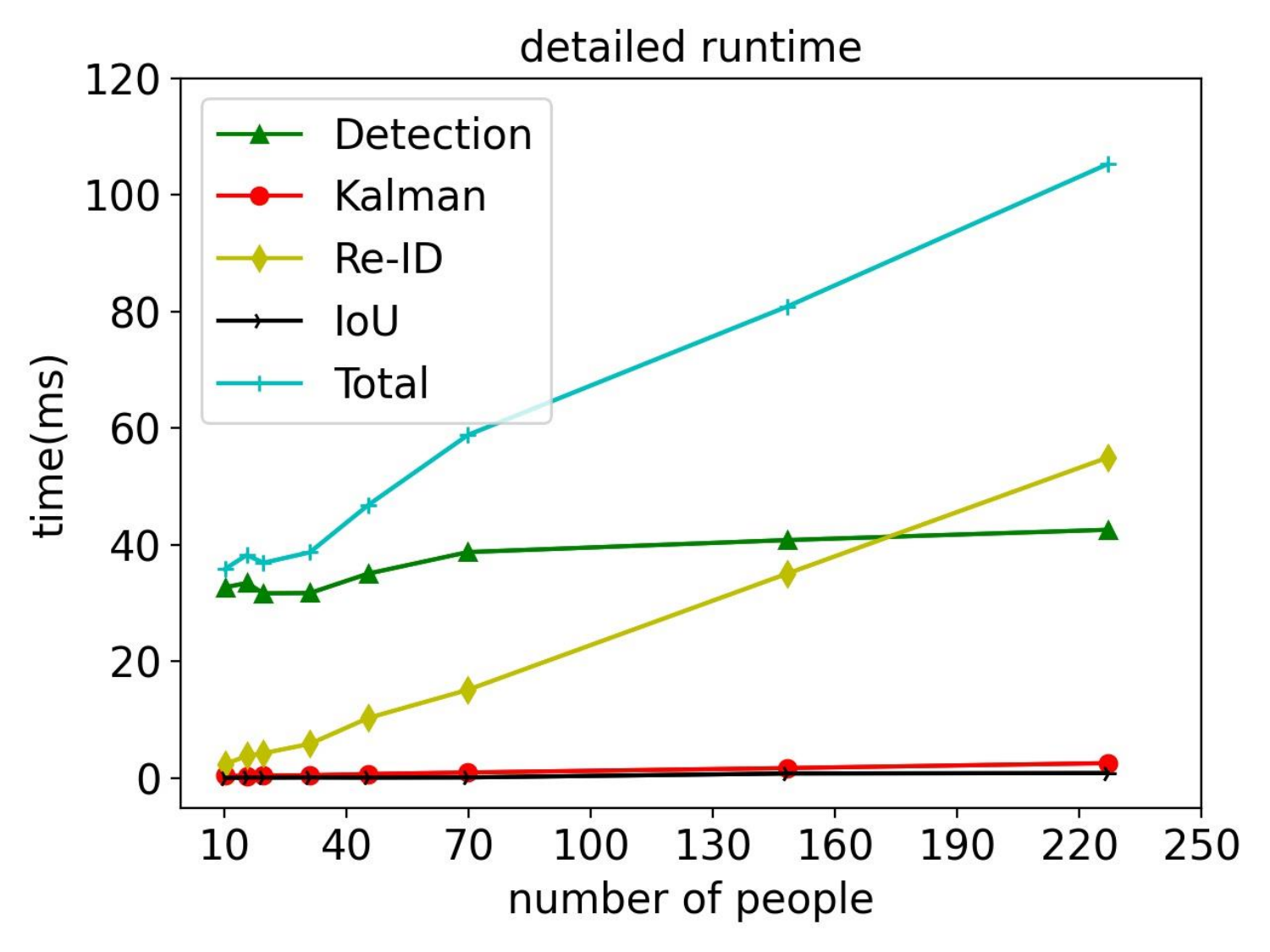}
\caption{Time spent on different parts of our whole MOT system. We run tracking on sequences with different density from the MOT17 dataset and the MOT20 dataset. }
\label{fig: time}
\end{figure}

We also present a detailed runtime breakdown of different components including detection, re-ID matching, Kalman Filter and IoU matching. We test runtime on sequences with different density (average number of pedestrians per frame). The results are shown in Fig \ref{fig: time}. The time spent on joint detection and re-ID is minimally affected by density. The time spent on Kalman Filter and IoU matching are around $1 ms$ or $2 ms$ and can be ignored. The time spent on re-ID matching increases linearly with the increase of density. This is because a large amount of time is cost on updating the re-ID feature of each tracklet.

\setlength{\tabcolsep}{3pt}
\begin{table}
\begin{center}
\caption{Evaluation of the three ingredients in the data association model. The backbone is DLA-34. The best results are shown in {\bf bold}.}
\label{table:association}
\begin{tabular}{cccccc}
\toprule
Box IoU & Re-ID Features & Kalman Filter & MOTA $\uparrow$ & IDF1 $\uparrow$ & IDs $\downarrow$\\
\midrule
$\checkmark$ &  &  & 67.8 & 67.2 & 648\\
 & $\checkmark$ &  & 68.1 & 70.3 & 435\\
 & $\checkmark$ & $\checkmark$ & 68.9 & 71.8 & 342\\
$\checkmark$ & $\checkmark$ & $\checkmark$ & \textbf{69.1} & \textbf{72.8} & \textbf{299}\\
\bottomrule
\end{tabular}
\end{center}
\end{table}
\setlength{\tabcolsep}{1.4pt}

\subsubsection{Visualization of Re-ID Similarity}
We use re-ID similarity maps to demonstrate the discriminative ability of re-ID features in Figure~\ref{fig:emb}. We randomly choose two frames from our validation set. The first frame contains the query instance and the second frame contains the target instance that has the same ID. We obtain the re-ID similarity maps by computing the cosine similarity between the re-ID feature of the query instance and the whole re-ID feature map of the target frame, as described in Section~\ref{sec:anchor} and Section~\ref{sec:backbone} respectively. By comparing the similarity maps of ResNet-34 and ResNet-34-det, we can see that training the re-ID branch is important. By comparing DLA-34 and ResNet-34, we can see that multi-layer feature aggregation can get more discriminative re-ID features. Among all the sampling strategies, the proposed Center and Center-BI can better discriminate the target object from surrounding objects in crowded scenes.

\subsection{Single Image Training}
We first pre-train \emph{FairMOT} on the CrowdHuman dataset \citep{shao2018crowdhuman}. In particular, we assign a unique identity label for each bounding box and train \emph{FairMOT} using the method described in section \ref{sec:training}. Then we finetune the pre-trained model on the target dataset MOT17.

Table~\ref{table:weakly} shows the results. First, the pre-trained model can be directly used as a tracker and get acceptable results on MOT datasets such as MOT17. This is because the CrowdHuman dataset can boost the human detection performance and also has strong domain generalization ability. Our training of the re-ID features further enhances the association ability of the tracker. Second, pre-training on CrowdHuman outperforms directly training on the MOT17 dataset by a large margin. Third, the single image training model even outperforms the model trained on the ``MIX'' and MOT17 datasets with identity annotations. The results validate the effectiveness of the proposed single image pre-training, which saves lots of annotation efforts and makes \emph{FairMOT} more attractive in real applications. 

\setlength{\tabcolsep}{5pt}
\begin{table}
\begin{center}
\caption{Effects of single image training on the validation set of MOT17. ``CH'' and ``MIX'' stand for CrowdHuman and the composed five datasets introduced in Section~\ref{sec:dataset}, respectively. \textsuperscript{*} means no identity annotations are used. The best results are shown in {\bf bold}.}
\label{table:weakly}
\begin{tabular}{lccccc}
\toprule
Training Data & MOTA $\uparrow$ & IDF1 $\uparrow$ & IDs $\downarrow$ & AP$\uparrow$ & TPR $\uparrow$\\
\midrule
CH\textsuperscript{*} & 64.1 & 64.9 & 476 & 80.5 & 79.9\\
MOT17 & 67.5 & 69.9 & 408 & 79.6 & 93.4\\
CH\textsuperscript{*}+MOT17 & \textbf{71.1} & \textbf{75.6} & 327 & \textbf{83.0} & 93.6\\
MIX+MOT17 & 69.1 & 72.8 & \textbf{299} & 81.2 & \textbf{94.4}\\
\bottomrule
\end{tabular}
\end{center}
\end{table}
\setlength{\tabcolsep}{1.4pt}

\setlength{\tabcolsep}{1.5pt}
\begin{table}[!htb]
\begin{center}
\caption{Comparison of the state-of-the-art one-shot trackers on the 2DMOT15 dataset. ``MIX'' represents the large scale training dataset and ``MOT17 Seg'' stands for the 4 videos with segmentation labels in the MOT17 dataset. The best results of the same training data are shown in {\bf bold}.}
\label{table:oneshot}
\begin{tabular}{llcccccccc}
\toprule
Training Data & Method & MOTA$\uparrow$ & IDF1$\uparrow$ & IDs$\downarrow$ & FP$\downarrow$ & FN$\downarrow$ & FPS$\uparrow$\\
\midrule
MIX & JDE & 67.5 & 66.7 & 218 & 1881 & \textbf{2083} & 26.0\\
 & FairMOT(ours) & \textbf{77.2} & \textbf{79.8} & \textbf{80} & \textbf{757} & 2094 & \textbf{30.9}\\
\midrule
MOT17 Seg & Track R-CNN & 69.2 & 49.4 & 294 & 1328 & {\bf 2349} & 2.0\\
 & FairMOT(ours) & {\bf 70.2} & {\bf 64.0} & {\bf 96} & {\bf 1209} & 2537 & {\bf 30.9}\\
\bottomrule
\end{tabular}
\end{center}
\end{table}
\setlength{\tabcolsep}{1pt}

\setlength{\tabcolsep}{10pt}
\begin{table*}[!htb]
\small
\begin{center}
\caption{Comparison of the state-of-the-art methods under the ``private detector'' protocol. It is noteworthy that FPS considers both detection and association time. The one-shot trackers are labeled by ``*''. The best results of each dataset are shown in {\bf bold}. }
\label{table:sota}
\begin{tabular}{llcccccc}
\toprule
Dataset & Tracker & MOTA$\uparrow$ & IDF1$\uparrow$ & MT$\uparrow$ & ML$\downarrow$ & IDs$\downarrow$ & FPS$\uparrow$\\
\midrule
MOT15 & MDP\_SubCNN\citep{xiang2015learning} & 47.5 & 55.7 & 30.0\% & 18.6\% & 628 & \textless 1.7\\
 & CDA\_DDAL\citep{bae2017confidence} & 51.3 & 54.1 & 36.3\% & 22.2\% & 544 & \textless 1.2\\
 & EAMTT\citep{sanchez2016online} & 53.0 & 54.0 & 35.9\% & 19.6\% & 7538 & \textless 4.0\\ 
 & AP\_HWDPL\citep{chen2017online} & 53.0 & 52.2 & 29.1\% & 20.2\% & 708 & 6.7\\
 & RAR15\citep{fang2018recurrent} & 56.5 & 61.3 & 45.1\% & 14.6\% & {\bf 428} & \textless 3.4\\
 & TubeTK\textsuperscript{*}\citep{pang2020tubetk} & 58.4 & 53.1 & 39.3\% & 18.0\% & 854 & 5.8\\
 & FairMOT (Ours)\textsuperscript{*} & {\bf 60.6} & {\bf 64.7} & {\bf 47.6\%} & {\bf 11.0\%} & 591 & {\bf 30.5}\\
\midrule
MOT16 & EAMTT\citep{sanchez2016online} & 52.5 & 53.3 & 19.9\% & 34.9\% & 910 & \textless 5.5\\
 & SORTwHPD16\citep{bewley2016simple} & 59.8 & 53.8 & 25.4\% & 22.7\% & 1423 & \textless 8.6\\
 & DeepSORT\_2\citep{wojke2017simple} & 61.4 & 62.2 & 32.8\% & 18.2\% & 781 & \textless 6.4\\
 & RAR16wVGG\citep{fang2018recurrent} & 63.0 & 63.8 & 39.9\% & 22.1\% & {\bf 482} & \textless 1.4\\
 & VMaxx\citep{wan2018multi} & 62.6 & 49.2 & 32.7\% & 21.1\% & 1389 & \textless 3.9\\
 & TubeTK\textsuperscript{*}\citep{pang2020tubetk} & 64.0 & 59.4 & 33.5\% & 19.4\% & 1117 & 1.0\\
 & JDE\textsuperscript{*}\citep{wang2020towards} & 64.4 & 55.8 & 35.4\% & 20.0\% & 1544 & 18.5\\
 & TAP\citep{zhou2018online} & 64.8 & {\bf 73.5} & 38.5\% & 21.6\% & 571 & \textless 8.0\\
 & CNNMTT\citep{mahmoudi2019multi} & 65.2 & 62.2 & 32.4\% & 21.3\% & 946 & \textless 5.3\\
 & POI\citep{yu2016poi} & 66.1 & 65.1 & 34.0\% & 20.8\% & 805 & \textless 5.0\\
 & CTrackerV1\textsuperscript{*}\citep{peng2020chained} & 67.6 & 57.2 & 32.9\% & 23.1\% & 1897 & 6.8\\
 & FairMOT (Ours)\textsuperscript{*} & {\bf 74.9} & 72.8 & {\bf 44.7\%} & {\bf 15.9\%} & 1074 & {\bf 25.9}\\
\midrule
MOT17 & SST\citep{sun2019deep} & 52.4 & 49.5 & 21.4\% & 30.7\% & 8431 & \textless 3.9\\
 & TubeTK\textsuperscript{*}\citep{pang2020tubetk} & 63.0 & 58.6 & 31.2\% & 19.9\% & 4137 & 3.0\\
 & CTrackerV1\textsuperscript{*}\citep{peng2020chained} & 66.6 & 57.4 & 32.2\% & 24.2\% & 5529 & 6.8\\
 & CenterTrack\textsuperscript{*}\citep{zhou2020tracking} & 67.8 & 64.7 & 34.6\% & 24.6\% & \textbf{2583} & 17.5\\
 & FairMOT (Ours)\textsuperscript{*} & {\bf 73.7} & {\bf 72.3} & {\bf 43.2\%} & {\bf 17.3\%} & 3303 & {\bf 25.9}\\
\midrule
MOT20 & FairMOT (Ours)\textsuperscript{*} & {\bf 61.8} & {\bf 67.3} & {\bf 68.8\%} & {\bf 7.6\%} & {\bf 5243} & {\bf 13.2}\\
\bottomrule
\end{tabular}
\end{center}
\end{table*}

\subsection{Results on MOTChallenge}

We compare our approach to the state-of-the-art (SOTA) methods including both the one-shot methods and the two-step methods. 

\subsubsection{Comparing with One-Shot SOTA MOT Methods}
There are two published works of JDE \citep{wang2020towards} and TrackRCNN \citep{voigtlaender2019mots} that jointly perform object detection and identity feature embedding. We compare our approach to both of them. Following the previous work \citep{wang2020towards}, the testing dataset contains 6 videos from 2DMOT15. \emph{FairMOT} uses the same training data as the two methods as described in their papers. In particular, when we compare to JDE, both \emph{FairMOT} and JDE use the large scale composed dataset described in Section~\ref{sec:dataset}. Since Track R-CNN requires segmentation labels to train the network, it only uses 4 videos of the MOT17 dataset which has segmentation labels as training data. In this case, we also use the 4 videos to train our model. The CLEAR metric \citep{bernardin2008evaluating} and IDF1 \citep{ristani2016performance} are used to measure their performance.

The results are shown in Table~\ref{table:oneshot}. We can see that our approach remarkably outperforms JDE \citep{wang2020towards}. In particular, the number of ID switches reduces from $218$ to $80$ which is big improvement in terms of user experience. The results validate the effectiveness of the \emph{anchor-free} approach over the previous \emph{anchor-based} one. The inference speed is near video rate for the both methods with ours being faster. Compared with Track R-CNN \citep{voigtlaender2019mots}, their detection results are slightly better than ours (with lower \emph{FN}). However, \emph{FairMOT} achieves much higher \emph{IDF1} score ($64.0$ vs. $49.4$) and fewer \emph{ID switches} ($96$ vs. $294$). This is mainly because Track R-CNN follows the ``detection first, re-ID secondary" framework and use anchors which also introduce ambiguity to the re-ID task.

\subsubsection{Comparing with Other SOTA MOT Methods}
We compare our approach to the state-of-the-art trackers including the two-step methods in Table~\ref{table:sota}. Since we do not use the public detection results, the ``private detector'' protocol is adopted. We report results on the testing sets of the 2DMOT15, MOT16, MOT17 and MOT20 datasets, respectively. Note that all of the results are directly obtained from the official MOT challenge evaluation server.

Our approach ranks first among all \emph{online} and \emph{offline} trackers on the four datasets. In particular, it outperforms other methods by a large margin. This is a very strong result especially considering that our approach is very simple. In addition, our approach achieves video rate inference. In contrast, most high-performance trackers such as \citep{fang2018recurrent,yu2016poi} are usually slower than ours. Our approach also ranks second under a very recent local MOT metric ALTA \citep{valmadre2021local}, which further indicates that our approach achieves very high tracking performance (Table~\ref{table:sota}).

\setlength{\tabcolsep}{2pt}
\begin{table}[!htbp]
\begin{center}
\caption{Results of the MOT17 test set when using different datasets for training. ``MIX'' represents the large scale dataset mentioned in part 4.1 and ``CH'' is short for the CrowdHuman dataset. All the results are obtained from the MOT challenge server. The best results are shown in {\bf bold}.}
\label{table:dataset}
\begin{tabular}{l|ccc|ccc}
\toprule
Training Data & Images & Boxes & Identities & MOTA$\uparrow$ & IDF1$\uparrow$ & IDs$\downarrow$ \\
\midrule
MOT17 & 5K & 112K & 0.5K & 69.8 & 69.9 & 3996\\
MOT17+MIX & 54K & 270K & 8.7K & 72.9 & \textbf{73.2} & 3345\\
MOT17+MIX+CH & 73K & 740K & 8.7K & \textbf{73.7} & 72.3 & \textbf{3303}\\
\bottomrule
\end{tabular}
\end{center}
\end{table}
\setlength{\tabcolsep}{1.4pt}

\subsubsection{Training Data Ablation Study}
We also evaluate the performance of \emph{FairMOT} using different amount of training data in Table~\ref{table:dataset}. We can achieve $69.8$ \emph{MOTA} when only using the MOT17 dataset for training, which already outperforms other methods using more training data. When we use the same training data as JDE \citep{wang2020towards}, we can achieve $72.9$ \emph{MOTA}, which remarkably outperforms JDE. In addition, when we perform single image training on the CrowdHuman dataset, the \emph{MOTA} score improves to $73.7$. The results suggest that our approach is not data hungry which is a big advantage in practical applications.

\begin{figure*}
	\centering
	\includegraphics[width=6.5in]{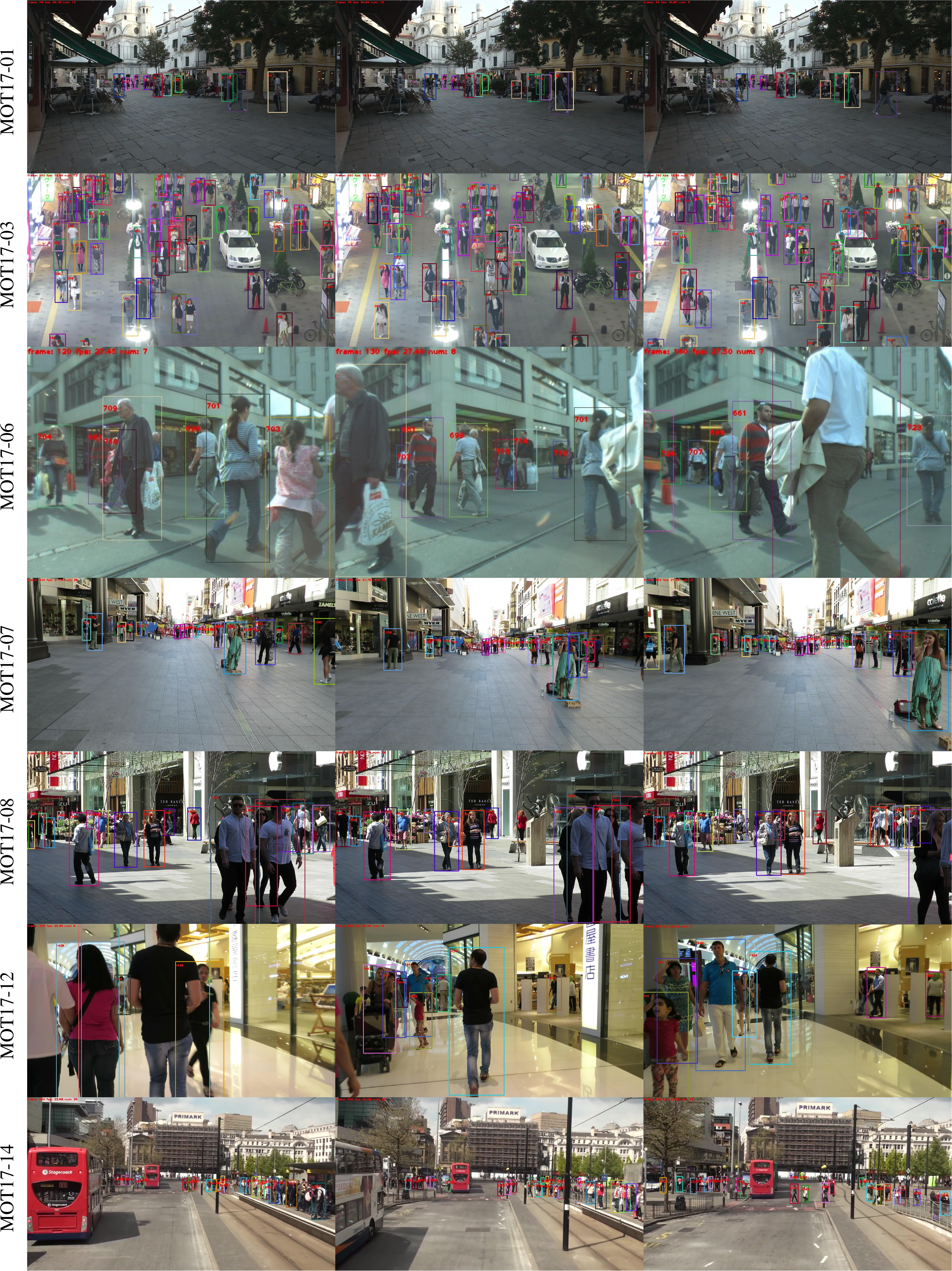}
	\caption{Example tracking results of our method on the test set of MOT17. Each row shows the results of sampled frames in chronological order of a video sequence. Bounding boxes and identities are marked in the images. Bounding boxes with different colors represent different identities. Best viewed in color.
	}
	\label{fig:tracking}
\end{figure*}

\subsection{Qualitative Results}
Figure~\ref{fig:tracking} visualizes several tracking results of \emph{FairMOT} on the test set of MOT17 \citep{milan2016mot16}. From the results of MOT17-01, we can see that our method can assign correct identities with the help of high-quality re-ID features when two pedestrians cross over each other. Trackers using bounding box IoUs \citep{bewley2016simple,bochinski2017high} usually cause identity switches under these circumstances. From the results of MOT17-03, we can see that our method perform well under crowded scenes. From the results of MOT17-08, we can see that our method can keep both correct identities and correct bounding boxes when the pedestrians are heavily occluded. The results of MOT17-06 and MOT17-12 show that our method can deal with large scale variations. This mainly attributes to the using of multi-layer feature aggregation. Our method can detect small objects accurately as shown in the results of MOT17-07 and MOT17-14. 

\section{Summary and Future Work}
Start from studying why the previous one-shot methods \citep{wang2020towards} fail to achieve comparable results as the two-step methods, we find that the use of anchors in object detection and identity embedding is the main reason for the degraded results. In particular, multiple nearby anchors, which correspond to different parts of an object, may be responsible for estimating the same identity which causes ambiguities for network training. Further, we find the feature unfairness issue and feature dimension issue between the detection and re-ID tasks in previous MOT frameworks. By addressing these problems in an anchor-free single-shot deep network, we propose \emph{FairMOT}. It outperforms the previous state-of-the-art methods on several benchmark datasets by a large margin in terms of both tracking accuracy and inference speed. Besides, FairMOT is inherently training data-efficient and we propose single image training of multi-object trackers only using bounding box annotated images, which both make our method more appealing in real applications \citep{zhang2021voxeltrack}.

\section*{Acknowledgements}
This work was in part supported by NSFC (No. 61733007 and No. 61876212) and MSRA Collaborative Research Fund. We thank all the anonymous reviewers for their valuable suggestions.

\bibliographystyle{spbasic}      
\bibliography{ref.bib}   

\begin{thebibliography}{107}
\providecommand{\natexlab}[1]{#1}
\providecommand{\url}[1]{{#1}}
\providecommand{\urlprefix}{URL }
\expandafter\ifx\csname urlstyle\endcsname\relax
  \providecommand{\doi}[1]{DOI~\discretionary{}{}{}#1}\else
  \providecommand{\doi}{DOI~\discretionary{}{}{}\begingroup
  \urlstyle{rm}\Url}\fi
\providecommand{\eprint}[2][]{\url{#2}}

\bibitem[{Bae and Yoon(2014)}]{bae2014robust}
Bae SH, Yoon KJ (2014) Robust online multi-object tracking based on tracklet
  confidence and online discriminative appearance learning. In: Proceedings of
  the IEEE conference on computer vision and pattern recognition, pp 1218--1225

\bibitem[{Bae and Yoon(2017)}]{bae2017confidence}
Bae SH, Yoon KJ (2017) Confidence-based data association and discriminative
  deep appearance learning for robust online multi-object tracking. IEEE
  transactions on pattern analysis and machine intelligence 40(3):595--610

\bibitem[{Berclaz et~al.(2011)Berclaz, Fleuret, Turetken, and
  Fua}]{berclaz2011multiple}
Berclaz J, Fleuret F, Turetken E, Fua P (2011) Multiple object tracking using
  k-shortest paths optimization. IEEE transactions on pattern analysis and
  machine intelligence 33(9):1806--1819

\bibitem[{Bergmann et~al.(2019)Bergmann, Meinhardt, and
  Leal-Taixe}]{bergmann2019tracking}
Bergmann P, Meinhardt T, Leal-Taixe L (2019) Tracking without bells and
  whistles. In: ICCV, pp 941--951

\bibitem[{Bernardin and Stiefelhagen(2008)}]{bernardin2008evaluating}
Bernardin K, Stiefelhagen R (2008) Evaluating multiple object tracking
  performance: the clear mot metrics. EURASIP Journal on Image and Video
  Processing 2008:1--10

\bibitem[{Bewley et~al.(2016)Bewley, Ge, Ott, Ramos, and
  Upcroft}]{bewley2016simple}
Bewley A, Ge Z, Ott L, Ramos F, Upcroft B (2016) Simple online and realtime
  tracking. In: ICIP, IEEE, pp 3464--3468

\bibitem[{Bochinski et~al.(2017)Bochinski, Eiselein, and
  Sikora}]{bochinski2017high}
Bochinski E, Eiselein V, Sikora T (2017) High-speed tracking-by-detection
  without using image information. In: 2017 14th IEEE International Conference
  on Advanced Video and Signal Based Surveillance (AVSS), IEEE, pp 1--6

\bibitem[{Bolme et~al.(2010)Bolme, Beveridge, Draper, and
  Lui}]{bolme2010visual}
Bolme DS, Beveridge JR, Draper BA, Lui YM (2010) Visual object tracking using
  adaptive correlation filters. In: CVPR, IEEE, pp 2544--2550

\bibitem[{Bras{\'o} and Leal-Taix{\'e}(2020)}]{braso2020learning}
Bras{\'o} G, Leal-Taix{\'e} L (2020) Learning a neural solver for multiple
  object tracking. In: Proceedings of the IEEE/CVF Conference on Computer
  Vision and Pattern Recognition, pp 6247--6257

\bibitem[{Cai and Vasconcelos(2018)}]{cai2018cascade}
Cai Z, Vasconcelos N (2018) Cascade r-cnn: Delving into high quality object
  detection. In: CVPR, pp 6154--6162

\bibitem[{Chao et~al.(2019)Chao, Kao, Ruan, Huang, and Lin}]{chao2019hardnet}
Chao P, Kao CY, Ruan YS, Huang CH, Lin YL (2019) Hardnet: A low memory traffic
  network. In: Proceedings of the IEEE/CVF International Conference on Computer
  Vision, pp 3552--3561

\bibitem[{Chen et~al.(2017)Chen, Ai, Shang, Zhuang, and Bai}]{chen2017online}
Chen L, Ai H, Shang C, Zhuang Z, Bai B (2017) Online multi-object tracking with
  convolutional neural networks. In: 2017 IEEE International Conference on
  Image Processing (ICIP), IEEE, pp 645--649

\bibitem[{Chen et~al.(2018{\natexlab{a}})Chen, Ai, Zhuang, and
  Shang}]{chen2018real}
Chen L, Ai H, Zhuang Z, Shang C (2018{\natexlab{a}}) Real-time multiple people
  tracking with deeply learned candidate selection and person
  re-identification. In: 2018 IEEE International Conference on Multimedia and
  Expo (ICME), IEEE, pp 1--6

\bibitem[{Chen et~al.(2018{\natexlab{b}})Chen, Badrinarayanan, Lee, and
  Rabinovich}]{chen2018gradnorm}
Chen Z, Badrinarayanan V, Lee CY, Rabinovich A (2018{\natexlab{b}}) Gradnorm:
  Gradient normalization for adaptive loss balancing in deep multitask
  networks. In: ICML, PMLR, pp 794--803

\bibitem[{Cheng et~al.(2020)Cheng, Xiao, Wang, Shi, Huang, and
  Zhang}]{cheng2020bottom}
Cheng B, Xiao B, Wang J, Shi H, Huang TS, Zhang L (2020) Higherhrnet:
  Scale-aware representation learning for bottom-up human pose estimation. In:
  CVPR

\bibitem[{Choi(2015)}]{choi2015near}
Choi W (2015) Near-online multi-target tracking with aggregated local flow
  descriptor. In: Proceedings of the IEEE international conference on computer
  vision, pp 3029--3037

\bibitem[{Chu and Ling(2019)}]{chu2019famnet}
Chu P, Ling H (2019) Famnet: Joint learning of feature, affinity and
  multi-dimensional assignment for online multiple object tracking. In: ICCV,
  pp 6172--6181

\bibitem[{Chu et~al.(2019)Chu, Fan, Tan, and Ling}]{chu2019online}
Chu P, Fan H, Tan CC, Ling H (2019) Online multi-object tracking with
  instance-aware tracker and dynamic model refreshment. In: 2019 IEEE Winter
  Conference on Applications of Computer Vision (WACV), IEEE, pp 161--170

\bibitem[{Dendorfer et~al.(2020)Dendorfer, Rezatofighi, Milan, Shi, Cremers,
  Reid, Roth, Schindler, and Leal-Taix{\'e}}]{dendorfer2020mot20}
Dendorfer P, Rezatofighi H, Milan A, Shi J, Cremers D, Reid I, Roth S,
  Schindler K, Leal-Taix{\'e} L (2020) Mot20: A benchmark for multi object
  tracking in crowded scenes. arXiv preprint arXiv:200309003

\bibitem[{Doll{\'a}r et~al.(2009)Doll{\'a}r, Wojek, Schiele, and
  Perona}]{dollar2009pedestrian}
Doll{\'a}r P, Wojek C, Schiele B, Perona P (2009) Pedestrian detection: A
  benchmark. In: CVPR, IEEE, pp 304--311

\bibitem[{Dong et~al.(2020)Dong, Li, Liao, Wang, Ren, and
  Qian}]{dong2020centripetalnet}
Dong Z, Li G, Liao Y, Wang F, Ren P, Qian C (2020) Centripetalnet: Pursuing
  high-quality keypoint pairs for object detection. In: CVPR, pp 10519--10528

\bibitem[{Duan et~al.(2019)Duan, Bai, Xie, Qi, Huang, and
  Tian}]{duan2019centernet}
Duan K, Bai S, Xie L, Qi H, Huang Q, Tian Q (2019) Centernet: Keypoint triplets
  for object detection. In: ICCV, pp 6569--6578

\bibitem[{Ess et~al.(2008)Ess, Leibe, Schindler, and Van~Gool}]{ess2008mobile}
Ess A, Leibe B, Schindler K, Van~Gool L (2008) A mobile vision system for
  robust multi-person tracking. In: CVPR, IEEE, pp 1--8

\bibitem[{Fang et~al.(2018)Fang, Xiang, Li, and Savarese}]{fang2018recurrent}
Fang K, Xiang Y, Li X, Savarese S (2018) Recurrent autoregressive networks for
  online multi-object tracking. In: WACV, IEEE, pp 466--475

\bibitem[{Feichtenhofer et~al.(2017)Feichtenhofer, Pinz, and
  Zisserman}]{feichtenhofer2017detect}
Feichtenhofer C, Pinz A, Zisserman A (2017) Detect to track and track to
  detect. In: Proceedings of the IEEE International Conference on Computer
  Vision, pp 3038--3046

\bibitem[{Felzenszwalb et~al.(2008)Felzenszwalb, McAllester, and
  Ramanan}]{felzenszwalb2008discriminatively}
Felzenszwalb P, McAllester D, Ramanan D (2008) A discriminatively trained,
  multiscale, deformable part model. In: CVPR, IEEE, pp 1--8

\bibitem[{Fu et~al.(2020)Fu, Zong, Li, Li, Yang, and Liu}]{fu2020model}
Fu J, Zong L, Li Y, Li K, Yang B, Liu X (2020) Model adaption object detection
  system for robot. In: 2020 39th Chinese Control Conference (CCC), IEEE, pp
  3659--3664

\bibitem[{Guo et~al.(2018)Guo, Haque, Huang, Yeung, and
  Fei-Fei}]{guo2018dynamic}
Guo M, Haque A, Huang DA, Yeung S, Fei-Fei L (2018) Dynamic task prioritization
  for multitask learning. In: Proceedings of the European Conference on
  Computer Vision (ECCV), pp 270--287

\bibitem[{Han et~al.(2020)Han, Huang, Wang, Yu, Liu, Pan, and
  Zhao}]{han2020mat}
Han S, Huang P, Wang H, Yu E, Liu D, Pan X, Zhao J (2020) Mat: Motion-aware
  multi-object tracking. arXiv preprint arXiv:200904794

\bibitem[{Han et~al.(2016)Han, Khorrami, Paine, Ramachandran, Babaeizadeh, Shi,
  Li, Yan, and Huang}]{han2016seq}
Han W, Khorrami P, Paine TL, Ramachandran P, Babaeizadeh M, Shi H, Li J, Yan S,
  Huang TS (2016) Seq-nms for video object detection. arXiv preprint
  arXiv:160208465

\bibitem[{He et~al.(2016)He, Zhang, Ren, and Sun}]{he2016deep}
He K, Zhang X, Ren S, Sun J (2016) Deep residual learning for image
  recognition. In: CVPR, pp 770--778

\bibitem[{He et~al.(2017)He, Gkioxari, Doll{\'a}r, and Girshick}]{he2017mask}
He K, Gkioxari G, Doll{\'a}r P, Girshick R (2017) Mask r-cnn. In: ICCV, pp
  2961--2969

\bibitem[{Henriques et~al.(2014)Henriques, Caseiro, Martins, and
  Batista}]{henriques2014high}
Henriques JF, Caseiro R, Martins P, Batista J (2014) High-speed tracking with
  kernelized correlation filters. IEEE transactions on pattern analysis and
  machine intelligence 37(3):583--596

\bibitem[{Henschel et~al.(2019)Henschel, Zou, and
  Rosenhahn}]{henschel2019multiple}
Henschel R, Zou Y, Rosenhahn B (2019) Multiple people tracking using body and
  joint detections. In: CVPRW, pp 0--0

\bibitem[{Hermans et~al.(2017)Hermans, Beyer, and Leibe}]{hermans2017defense}
Hermans A, Beyer L, Leibe B (2017) In defense of the triplet loss for person
  re-identification. arXiv preprint arXiv:170307737

\bibitem[{Hornakova et~al.(2020)Hornakova, Henschel, Rosenhahn, and
  Swoboda}]{hornakova2020lifted}
Hornakova A, Henschel R, Rosenhahn B, Swoboda P (2020) Lifted disjoint paths
  with application in multiple object tracking. In: International Conference on
  Machine Learning, PMLR, pp 4364--4375

\bibitem[{Kalman(1960)}]{kalman1960new}
Kalman RE (1960) A new approach to linear filtering and prediction problems. J
  Fluids Eng 82(1):35--45

\bibitem[{Kang et~al.(2016)Kang, Ouyang, Li, and Wang}]{kang2016object}
Kang K, Ouyang W, Li H, Wang X (2016) Object detection from video tubelets with
  convolutional neural networks. In: Proceedings of the IEEE conference on
  computer vision and pattern recognition, pp 817--825

\bibitem[{Kang et~al.(2017)Kang, Li, Xiao, Ouyang, Yan, Liu, and
  Wang}]{kang2017object}
Kang K, Li H, Xiao T, Ouyang W, Yan J, Liu X, Wang X (2017) Object detection in
  videos with tubelet proposal networks. In: Proceedings of the IEEE Conference
  on Computer Vision and Pattern Recognition, pp 727--735

\bibitem[{Kendall et~al.(2018)Kendall, Gal, and Cipolla}]{kendall2018multi}
Kendall A, Gal Y, Cipolla R (2018) Multi-task learning using uncertainty to
  weigh losses for scene geometry and semantics. In: CVPR, pp 7482--7491

\bibitem[{Kingma and Ba(2014)}]{kingma2014adam}
Kingma DP, Ba J (2014) Adam: A method for stochastic optimization. arXiv
  preprint arXiv:14126980

\bibitem[{Kokkinos(2017)}]{kokkinos2017ubernet}
Kokkinos I (2017) Ubernet: Training a universal convolutional neural network
  for low-, mid-, and high-level vision using diverse datasets and limited
  memory. In: CVPR, pp 6129--6138

\bibitem[{Kuhn(1955)}]{kuhn1955hungarian}
Kuhn HW (1955) The hungarian method for the assignment problem. Naval research
  logistics quarterly 2(1-2):83--97

\bibitem[{Law and Deng(2018)}]{law2018cornernet}
Law H, Deng J (2018) Cornernet: Detecting objects as paired keypoints. In:
  ECCV, pp 734--750

\bibitem[{Leal-Taix{\'e} et~al.(2015)Leal-Taix{\'e}, Milan, Reid, Roth, and
  Schindler}]{leal2015motchallenge}
Leal-Taix{\'e} L, Milan A, Reid I, Roth S, Schindler K (2015) Motchallenge
  2015: Towards a benchmark for multi-target tracking. arXiv preprint
  arXiv:150401942

\bibitem[{Liang et~al.(2020)Liang, Zhang, Lu, Zhou, Li, Ye, and
  Zou}]{liang2020rethinking}
Liang C, Zhang Z, Lu Y, Zhou X, Li B, Ye X, Zou J (2020) Rethinking the
  competition between detection and reid in multi-object tracking. arXiv
  preprint arXiv:201012138

\bibitem[{Lin et~al.(2014)Lin, Maire, Belongie, Hays, Perona, Ramanan,
  Doll{\'a}r, and Zitnick}]{lin2014microsoft}
Lin TY, Maire M, Belongie S, Hays J, Perona P, Ramanan D, Doll{\'a}r P, Zitnick
  CL (2014) Microsoft coco: Common objects in context. In: ECCV, Springer, pp
  740--755

\bibitem[{Lin et~al.(2017{\natexlab{a}})Lin, Doll{\'a}r, Girshick, He,
  Hariharan, and Belongie}]{lin2017feature}
Lin TY, Doll{\'a}r P, Girshick R, He K, Hariharan B, Belongie S
  (2017{\natexlab{a}}) Feature pyramid networks for object detection. In: CVPR,
  pp 2117--2125

\bibitem[{Lin et~al.(2017{\natexlab{b}})Lin, Goyal, Girshick, He, and
  Doll{\'a}r}]{lin2017focal}
Lin TY, Goyal P, Girshick R, He K, Doll{\'a}r P (2017{\natexlab{b}}) Focal loss
  for dense object detection. In: ICCV, pp 2980--2988

\bibitem[{Liu et~al.(2019)Liu, Johns, and Davison}]{liu2019end}
Liu S, Johns E, Davison AJ (2019) End-to-end multi-task learning with
  attention. In: Proceedings of the IEEE/CVF Conference on Computer Vision and
  Pattern Recognition, pp 1871--1880

\bibitem[{Lu et~al.(2020)Lu, Rathod, Votel, and Huang}]{lu2020retinatrack}
Lu Z, Rathod V, Votel R, Huang J (2020) Retinatrack: Online single stage joint
  detection and tracking. In: Proceedings of the IEEE/CVF conference on
  computer vision and pattern recognition, pp 14668--14678

\bibitem[{Luo et~al.(2017)Luo, Ma, Wang, and Wang}]{luo2017learning}
Luo C, Ma C, Wang C, Wang Y (2017) Learning discriminative activated simplices
  for action recognition. In: AAAI

\bibitem[{Luo et~al.(2019{\natexlab{a}})Luo, Gu, Liao, Lai, and
  Jiang}]{luo2019bag}
Luo H, Gu Y, Liao X, Lai S, Jiang W (2019{\natexlab{a}}) Bag of tricks and a
  strong baseline for deep person re-identification. In: Proceedings of the
  IEEE/CVF Conference on Computer Vision and Pattern Recognition Workshops, pp
  0--0

\bibitem[{Luo et~al.(2019{\natexlab{b}})Luo, Xie, Wang, and
  Zeng}]{luo2019detect}
Luo H, Xie W, Wang X, Zeng W (2019{\natexlab{b}}) Detect or track: Towards
  cost-effective video object detection/tracking. In: Proceedings of the AAAI
  Conference on Artificial Intelligence, vol~33, pp 8803--8810

\bibitem[{Mahmoudi et~al.(2019)Mahmoudi, Ahadi, and
  Rahmati}]{mahmoudi2019multi}
Mahmoudi N, Ahadi SM, Rahmati M (2019) Multi-target tracking using cnn-based
  features: Cnnmtt. Multimedia Tools and Applications 78(6):7077--7096

\bibitem[{Milan et~al.(2013)Milan, Roth, and Schindler}]{milan2013continuous}
Milan A, Roth S, Schindler K (2013) Continuous energy minimization for
  multitarget tracking. IEEE transactions on pattern analysis and machine
  intelligence 36(1):58--72

\bibitem[{Milan et~al.(2016)Milan, Leal-Taix{\'e}, Reid, Roth, and
  Schindler}]{milan2016mot16}
Milan A, Leal-Taix{\'e} L, Reid I, Roth S, Schindler K (2016) Mot16: A
  benchmark for multi-object tracking. arXiv preprint arXiv:160300831

\bibitem[{Pang et~al.(2020)Pang, Li, Zhang, Li, and Lu}]{pang2020tubetk}
Pang B, Li Y, Zhang Y, Li M, Lu C (2020) Tubetk: Adopting tubes to track
  multi-object in a one-step training model. In: Proceedings of the IEEE/CVF
  Conference on Computer Vision and Pattern Recognition, pp 6308--6318

\bibitem[{Pang et~al.(2021)Pang, Qiu, Li, Chen, Li, Darrell, and
  Yu}]{pang2021quasi}
Pang J, Qiu L, Li X, Chen H, Li Q, Darrell T, Yu F (2021) Quasi-dense
  similarity learning for multiple object tracking. In: Proceedings of the
  IEEE/CVF Conference on Computer Vision and Pattern Recognition, pp 164--173

\bibitem[{Peng et~al.(2020)Peng, Wang, Wan, Wu, Wang, Tai, Wang, Li, Huang, and
  Fu}]{peng2020chained}
Peng J, Wang C, Wan F, Wu Y, Wang Y, Tai Y, Wang C, Li J, Huang F, Fu Y (2020)
  Chained-tracker: Chaining paired attentive regression results for end-to-end
  joint multiple-object detection and tracking. In: European Conference on
  Computer Vision, Springer, pp 145--161

\bibitem[{Radosavovic et~al.(2020)Radosavovic, Kosaraju, Girshick, He, and
  Doll{\'a}r}]{radosavovic2020designing}
Radosavovic I, Kosaraju RP, Girshick R, He K, Doll{\'a}r P (2020) Designing
  network design spaces. In: Proceedings of the IEEE/CVF Conference on Computer
  Vision and Pattern Recognition, pp 10428--10436

\bibitem[{Ranjan et~al.(2017)Ranjan, Patel, and
  Chellappa}]{ranjan2017hyperface}
Ranjan R, Patel VM, Chellappa R (2017) Hyperface: A deep multi-task learning
  framework for face detection, landmark localization, pose estimation, and
  gender recognition. T-PAMI 41(1):121--135

\bibitem[{Redmon and Farhadi(2018)}]{redmon2018yolov3}
Redmon J, Farhadi A (2018) Yolov3: An incremental improvement. arXiv preprint
  arXiv:180402767

\bibitem[{Ren et~al.(2015)Ren, He, Girshick, and Sun}]{ren2015faster}
Ren S, He K, Girshick R, Sun J (2015) Faster r-cnn: Towards real-time object
  detection with region proposal networks. In: Advances in neural information
  processing systems, pp 91--99

\bibitem[{Ristani et~al.(2016)Ristani, Solera, Zou, Cucchiara, and
  Tomasi}]{ristani2016performance}
Ristani E, Solera F, Zou R, Cucchiara R, Tomasi C (2016) Performance measures
  and a data set for multi-target, multi-camera tracking. In: ECCV, Springer,
  pp 17--35

\bibitem[{Russakovsky et~al.(2015)Russakovsky, Deng, Su, Krause, Satheesh, Ma,
  Huang, Karpathy, Khosla, Bernstein et~al.}]{russakovsky2015imagenet}
Russakovsky O, Deng J, Su H, Krause J, Satheesh S, Ma S, Huang Z, Karpathy A,
  Khosla A, Bernstein M, et~al. (2015) Imagenet large scale visual recognition
  challenge. International journal of computer vision 115(3):211--252

\bibitem[{Sadeghian et~al.(2017)Sadeghian, Alahi, and
  Savarese}]{sadeghian2017tracking}
Sadeghian A, Alahi A, Savarese S (2017) Tracking the untrackable: Learning to
  track multiple cues with long-term dependencies. In: Proceedings of the IEEE
  International Conference on Computer Vision, pp 300--311

\bibitem[{Sanchez-Matilla et~al.(2016)Sanchez-Matilla, Poiesi, and
  Cavallaro}]{sanchez2016online}
Sanchez-Matilla R, Poiesi F, Cavallaro A (2016) Online multi-target tracking
  with strong and weak detections. In: ECCV, Springer, pp 84--99

\bibitem[{Sener and Koltun(2018)}]{sener2018multi}
Sener O, Koltun V (2018) Multi-task learning as multi-objective optimization.
  In: NIPS, pp 527--538

\bibitem[{Shan et~al.(2020)Shan, Wei, Deng, Huang, Hua, Cheng, and
  Liang}]{shan2020fgagt}
Shan C, Wei C, Deng B, Huang J, Hua XS, Cheng X, Liang K (2020) Fgagt:
  Flow-guided adaptive graph tracking. arXiv preprint arXiv:201009015

\bibitem[{Shao et~al.(2018)Shao, Zhao, Li, Xiao, Yu, Zhang, and
  Sun}]{shao2018crowdhuman}
Shao S, Zhao Z, Li B, Xiao T, Yu G, Zhang X, Sun J (2018) Crowdhuman: A
  benchmark for detecting human in a crowd. arXiv preprint arXiv:180500123

\bibitem[{Simonyan and Zisserman(2014)}]{simonyan2014very}
Simonyan K, Zisserman A (2014) Very deep convolutional networks for large-scale
  image recognition. arXiv preprint arXiv:14091556

\bibitem[{Sun et~al.(2020)Sun, Jiang, Zhang, Xie, Cao, Hu, Kong, Yuan, Wang,
  and Luo}]{sun2020transtrack}
Sun P, Jiang Y, Zhang R, Xie E, Cao J, Hu X, Kong T, Yuan Z, Wang C, Luo P
  (2020) Transtrack: Multiple-object tracking with transformer. arXiv preprint
  arXiv:201215460

\bibitem[{Sun et~al.(2021{\natexlab{a}})Sun, Jiang, Xie, Shao, Yuan, Wang, and
  Luo}]{peize2020onenet}
Sun P, Jiang Y, Xie E, Shao W, Yuan Z, Wang C, Luo P (2021{\natexlab{a}}) What
  makes for end-to-end object detection? In: Proceedings of the 38th
  International Conference on Machine Learning, PMLR, Proceedings of Machine
  Learning Research, vol 139, pp 9934--9944

\bibitem[{Sun et~al.(2021{\natexlab{b}})Sun, Zhang, Jiang, Kong, Xu, Zhan,
  Tomizuka, Li, Yuan, Wang et~al.}]{sun2021sparse}
Sun P, Zhang R, Jiang Y, Kong T, Xu C, Zhan W, Tomizuka M, Li L, Yuan Z, Wang
  C, et~al. (2021{\natexlab{b}}) Sparse r-cnn: End-to-end object detection with
  learnable proposals. In: Proceedings of the IEEE/CVF Conference on Computer
  Vision and Pattern Recognition, pp 14454--14463

\bibitem[{Sun et~al.(2019)Sun, Akhtar, Song, Mian, and Shah}]{sun2019deep}
Sun S, Akhtar N, Song H, Mian AS, Shah M (2019) Deep affinity network for
  multiple object tracking. IEEE transactions on pattern analysis and machine
  intelligence

\bibitem[{Tang et~al.(2019)Tang, Wang, Wang, Liu, Zeng, and
  Wang}]{tang2019object}
Tang P, Wang C, Wang X, Liu W, Zeng W, Wang J (2019) Object detection in videos
  by high quality object linking. IEEE transactions on pattern analysis and
  machine intelligence 42(5):1272--1278

\bibitem[{Tang et~al.(2017)Tang, Andriluka, Andres, and
  Schiele}]{tang2017multiple}
Tang S, Andriluka M, Andres B, Schiele B (2017) Multiple people tracking by
  lifted multicut and person re-identification. In: Proceedings of the IEEE
  Conference on Computer Vision and Pattern Recognition, pp 3539--3548

\bibitem[{Valmadre et~al.(2021)Valmadre, Bewley, Huang, Sun, Sminchisescu, and
  Schmid}]{valmadre2021local}
Valmadre J, Bewley A, Huang J, Sun C, Sminchisescu C, Schmid C (2021) Local
  metrics for multi-object tracking. arXiv preprint arXiv:210402631

\bibitem[{Vandenhende et~al.(2021)Vandenhende, Georgoulis, Van~Gansbeke,
  Proesmans, Dai, and Van~Gool}]{vandenhende2021multi}
Vandenhende S, Georgoulis S, Van~Gansbeke W, Proesmans M, Dai D, Van~Gool L
  (2021) Multi-task learning for dense prediction tasks: A survey. IEEE
  Transactions on Pattern Analysis and Machine Intelligence

\bibitem[{Voigtlaender et~al.(2019)Voigtlaender, Krause, Osep, Luiten, Sekar,
  Geiger, and Leibe}]{voigtlaender2019mots}
Voigtlaender P, Krause M, Osep A, Luiten J, Sekar BBG, Geiger A, Leibe B (2019)
  Mots: Multi-object tracking and segmentation. In: CVPR, pp 7942--7951

\bibitem[{Wan et~al.(2018)Wan, Wang, Kong, Zhao, and Deng}]{wan2018multi}
Wan X, Wang J, Kong Z, Zhao Q, Deng S (2018) Multi-object tracking using online
  metric learning with long short-term memory. In: 2018 25th IEEE International
  Conference on Image Processing (ICIP), IEEE, pp 788--792

\bibitem[{Wang et~al.(2013)Wang, Wang, and Yuille}]{wang2013approach}
Wang C, Wang Y, Yuille AL (2013) An approach to pose-based action recognition.
  In: CVPR, pp 915--922

\bibitem[{Wang et~al.(2020{\natexlab{a}})Wang, Sun, Cheng, Jiang, Deng, Zhao,
  Liu, Mu, Tan, Wang et~al.}]{wang2020deep}
Wang J, Sun K, Cheng T, Jiang B, Deng C, Zhao Y, Liu D, Mu Y, Tan M, Wang X,
  et~al. (2020{\natexlab{a}}) Deep high-resolution representation learning for
  visual recognition. IEEE transactions on pattern analysis and machine
  intelligence

\bibitem[{Wang et~al.(2020{\natexlab{b}})Wang, Zheng, Liu, Li, and
  Wang}]{wang2020towards}
Wang Z, Zheng L, Liu Y, Li Y, Wang S (2020{\natexlab{b}}) Towards real-time
  multi-object tracking. In: Computer Vision--ECCV 2020: 16th European
  Conference, Glasgow, UK, August 23--28, 2020, Proceedings, Part XI 16,
  Springer, pp 107--122

\bibitem[{Wen et~al.(2014)Wen, Li, Yan, Lei, Yi, and Li}]{wen2014multiple}
Wen L, Li W, Yan J, Lei Z, Yi D, Li SZ (2014) Multiple target tracking based on
  undirected hierarchical relation hypergraph. In: Proceedings of the IEEE
  conference on computer vision and pattern recognition, pp 1282--1289

\bibitem[{Wojke et~al.(2017)Wojke, Bewley, and Paulus}]{wojke2017simple}
Wojke N, Bewley A, Paulus D (2017) Simple online and realtime tracking with a
  deep association metric. In: 2017 IEEE international conference on image
  processing (ICIP), IEEE, pp 3645--3649

\bibitem[{Xiang et~al.(2015)Xiang, Alahi, and Savarese}]{xiang2015learning}
Xiang Y, Alahi A, Savarese S (2015) Learning to track: Online multi-object
  tracking by decision making. In: ICCV, pp 4705--4713

\bibitem[{Xiao et~al.(2017)Xiao, Li, Wang, Lin, and Wang}]{xiao2017joint}
Xiao T, Li S, Wang B, Lin L, Wang X (2017) Joint detection and identification
  feature learning for person search. In: CVPR, pp 3415--3424

\bibitem[{Xu et~al.(2019)Xu, Cao, Zhang, and Hu}]{xu2019spatial}
Xu J, Cao Y, Zhang Z, Hu H (2019) Spatial-temporal relation networks for
  multi-object tracking. In: Proceedings of the IEEE/CVF International
  Conference on Computer Vision, pp 3988--3998

\bibitem[{Yang et~al.(2016)Yang, Choi, and Lin}]{yang2016exploit}
Yang F, Choi W, Lin Y (2016) Exploit all the layers: Fast and accurate cnn
  object detector with scale dependent pooling and cascaded rejection
  classifiers. In: Proceedings of the IEEE conference on computer vision and
  pattern recognition, pp 2129--2137

\bibitem[{Yang et~al.(2019)Yang, Liu, Hu, Wang, and Lin}]{yang2019reppoints}
Yang Z, Liu S, Hu H, Wang L, Lin S (2019) Reppoints: Point set representation
  for object detection. In: ICCV, pp 9657--9666

\bibitem[{Yu et~al.(2016)Yu, Li, Li, Liu, Shi, and Yan}]{yu2016poi}
Yu F, Li W, Li Q, Liu Y, Shi X, Yan J (2016) Poi: Multiple object tracking with
  high performance detection and appearance feature. In: ECCV, Springer, pp
  36--42

\bibitem[{Yu et~al.(2018)Yu, Wang, Shelhamer, and Darrell}]{yu2018deep}
Yu F, Wang D, Shelhamer E, Darrell T (2018) Deep layer aggregation. In: CVPR,
  pp 2403--2412

\bibitem[{Zamir et~al.(2012)Zamir, Dehghan, and Shah}]{zamir2012gmcp}
Zamir AR, Dehghan A, Shah M (2012) Gmcp-tracker: Global multi-object tracking
  using generalized minimum clique graphs. In: European Conference on Computer
  Vision, Springer, pp 343--356

\bibitem[{Zhang et~al.(2008)Zhang, Li, and Nevatia}]{zhang2008global}
Zhang L, Li Y, Nevatia R (2008) Global data association for multi-object
  tracking using network flows. In: 2008 IEEE Conference on Computer Vision and
  Pattern Recognition, IEEE, pp 1--8

\bibitem[{Zhang et~al.(2017)Zhang, Benenson, and
  Schiele}]{zhang2017citypersons}
Zhang S, Benenson R, Schiele B (2017) Citypersons: A diverse dataset for
  pedestrian detection. In: CVPR, pp 3213--3221

\bibitem[{Zhang et~al.(2020)Zhang, Sheng, Wu, Wang, Lyu, Ke, and
  Xiong}]{zhang2020long}
Zhang Y, Sheng H, Wu Y, Wang S, Lyu W, Ke W, Xiong Z (2020) Long-term tracking
  with deep tracklet association. IEEE Transactions on Image Processing
  29:6694--6706

\bibitem[{Zhang et~al.(2021{\natexlab{a}})Zhang, Sun, Jiang, Yu, Yuan, Luo,
  Liu, and Wang}]{zhang2021bytetrack}
Zhang Y, Sun P, Jiang Y, Yu D, Yuan Z, Luo P, Liu W, Wang X
  (2021{\natexlab{a}}) Bytetrack: Multi-object tracking by associating every
  detection box. arXiv preprint arXiv:211006864

\bibitem[{Zhang et~al.(2021{\natexlab{b}})Zhang, Wang, Wang, Liu, and
  Zeng}]{zhang2021voxeltrack}
Zhang Y, Wang C, Wang X, Liu W, Zeng W (2021{\natexlab{b}}) Voxeltrack:
  Multi-person 3d human pose estimation and tracking in the wild. arXiv
  preprint arXiv:210802452

\bibitem[{Zheng et~al.(2017{\natexlab{a}})Zheng, Zhang, Sun, Chandraker, Yang,
  and Tian}]{zheng2017person}
Zheng L, Zhang H, Sun S, Chandraker M, Yang Y, Tian Q (2017{\natexlab{a}})
  Person re-identification in the wild. In: CVPR, pp 1367--1376

\bibitem[{Zheng et~al.(2017{\natexlab{b}})Zheng, Zheng, and
  Yang}]{zheng2017discriminatively}
Zheng Z, Zheng L, Yang Y (2017{\natexlab{b}}) A discriminatively learned cnn
  embedding for person reidentification. ACM Transactions on Multimedia
  Computing, Communications, and Applications (TOMM) 14(1):1--20

\bibitem[{Zhou et~al.(2019{\natexlab{a}})Zhou, Wang, and
  Kr{\"a}henb{\"u}hl}]{zhou2019objects}
Zhou X, Wang D, Kr{\"a}henb{\"u}hl P (2019{\natexlab{a}}) Objects as points.
  arXiv preprint arXiv:190407850

\bibitem[{Zhou et~al.(2019{\natexlab{b}})Zhou, Zhuo, and
  Krahenbuhl}]{zhou2019bottom}
Zhou X, Zhuo J, Krahenbuhl P (2019{\natexlab{b}}) Bottom-up object detection by
  grouping extreme and center points. In: CVPR, pp 850--859

\bibitem[{Zhou et~al.(2020)Zhou, Koltun, and
  Kr{\"a}henb{\"u}hl}]{zhou2020tracking}
Zhou X, Koltun V, Kr{\"a}henb{\"u}hl P (2020) Tracking objects as points. In:
  European Conference on Computer Vision, Springer, pp 474--490

\bibitem[{Zhou et~al.(2018)Zhou, Xing, Zhang, and Hu}]{zhou2018online}
Zhou Z, Xing J, Zhang M, Hu W (2018) Online multi-target tracking with
  tensor-based high-order graph matching. In: 2018 24th International
  Conference on Pattern Recognition (ICPR), IEEE, pp 1809--1814

\bibitem[{Zhu et~al.(2018)Zhu, Yang, Liu, Kim, Zhang, and Yang}]{zhu2018online}
Zhu J, Yang H, Liu N, Kim M, Zhang W, Yang MH (2018) Online multi-object
  tracking with dual matching attention networks. In: Proceedings of the
  European Conference on Computer Vision (ECCV), pp 366--382

\end{thebibliography}
\end{document}